\documentclass{article}
\PassOptionsToPackage{table}{xcolor}
\usepackage{xcolor}  

\usepackage[utf8]{inputenc}
\usepackage[T1]{fontenc}

\usepackage{neurips_2025}

\usepackage{svg}
\usepackage{float}

\usepackage[table]{xcolor}
\usepackage{array}

\usepackage{siunitx}
\usepackage{url,booktabs,amsfonts,nicefrac,microtype,xcolor}
\usepackage{hyperref}
\hypersetup{
    colorlinks=true,
    linkcolor=blue,
    filecolor=magenta,
    urlcolor=cyan,
    citecolor=blue, 
}

\usepackage{array}
\usepackage{amssymb}    
\usepackage{graphicx}   
\usepackage{makecell}

\newcolumntype{L}[1]{>{\raggedright\arraybackslash}p{#1}}
\newcolumntype{C}[1]{>{\centering\arraybackslash}p{#1}}

\newcommand{\yesmark}{{\color{green!60!black}\checkmark}}
\newcommand{\nomark}{{\color{red}$\times$}}
\newcommand{\somewhatmark}{{\color{orange}$\approx$}}

\usepackage{natbib}
\setcitestyle{authoryear,open={(},close={)}} 

\title{Context Is Not Comprehension}
\author{%
  Alex Pan\\
  Founder, Edictive\\
  \texttt{z5060939@zmail.unsw.edu.au}
  \and
  Mary-Anne Williams\\
  UNSW Business School\\
  UNSW AI Institute\\
  \texttt{mary-anne.williams@unsw.edu.au}
}
\date{} 

\begin{document}
\maketitle
\begin{abstract}
	The dominant way of judging Large Language Models (LLMs) has been to ask how well they can recall explicit facts from very long inputs. While today's best models achieve near-perfect recall, this masks a harder skill: performing multi-step reasoning and tracking intermediate state that never appears verbatim. We introduce Verbose ListOps (VLO), a benchmark that embeds deterministic ListOps computations inside narrative camouflage and, crucially, allows step-level evaluation of every intermediate result. Experiments show that models which solve raw ListOps with $\approx100\%$accuracy collapse on VLO after only 10 k tokens. By exposing where a model's reasoning chain first diverges, VLO moves assessment beyond sheer context length and toward genuine comprehension. VLO's generation pipeline is task-agnostic: it can weave any deterministically verifiable reasoning schema—arithmetic, symbolic, abductive, inductive or defeasible—into narrative form. This makes VLO a reusable test-bed for the next wave of reasoning-centric model designs, not merely those with step-explicit scaffolds.
\end{abstract}

\section{Introduction}
Despite million-token context windows, today's LLMs still stumble on tasks that a human finds trivial: following a multi-step argument buried inside a distracting story \citep{liu-etal-2024-modeling, shi2023large, an2023l}. This ability—to ignore irrelevant detail, maintain intermediate conclusions, and synthesise them later—is the backbone of knowledge work, from a lawyer interpreting clauses to a sales manager inferring intent from chatty transcripts. Yet benchmarks rarely measure it. To bridge this gap, we introduce Verbose ListOps (VLO), a novel benchmark that isolates this ability in a controlled setting, challenging models to reason, not just locate information.

Existing long-context suites chiefly test (i) factual recall in "needle-in-a-haystack" settings \citep{kamradt2023needle,li_mo2024needlebench} or (ii) shallow reading comprehension \citep{an2023l, zhang2024infinitebench_acl}. They do not control reasoning depth, nor do they hide intermediate answers, so models can succeed through pattern matching rather than computation.

Verbose ListOps (VLO) fills this gap. ListOps is a synthetic benchmark where models perform nested arithmetic operations on sequences of integers following a tree structure—for example, \texttt{[MAX 3 7 [MIN 1 9 5] 2]} requires computing the inner \texttt{MIN} first, then finding the maximum among all values. We convert each ListOps node into a short story fragment, hide every intermediate value behind a 'narrative anchor', and add semantically related distractors. The resulting benchmark forces a model to compute, cache, and reuse internal state.

\begin{figure}[H]
	\centering
	\includegraphics[width=1\textwidth]{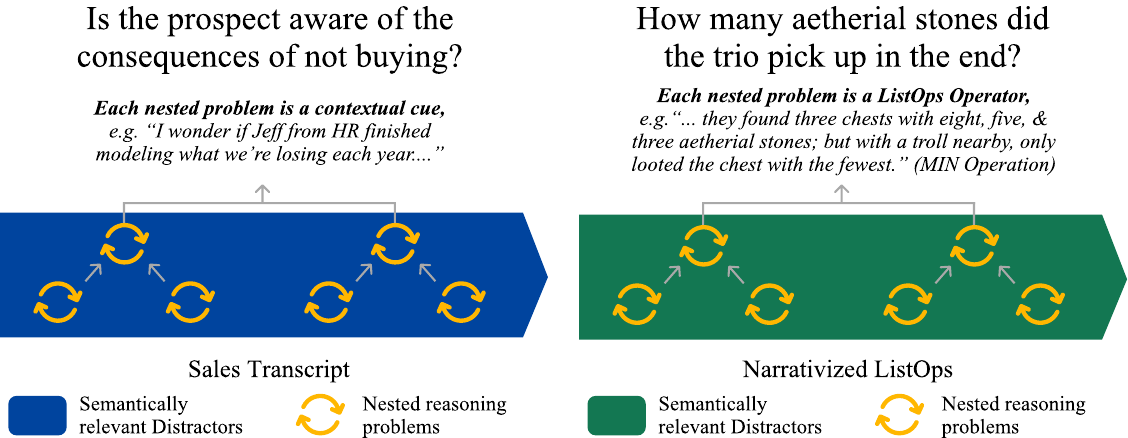}
	\caption{VLO embeds ListOps problems in a narrative, forcing LLMs to find and solve each subproblem, track intermediate states, and ignore distractors for the final answer, emulating complex human text synthesis. \textbf{Takeaway:} VLO reveals LLMs needs more than just long-context processing—they must compute within distracting narratives where intermediate results remain implicit.}
	\label{fig:ListOps_customization}
\end{figure}

In this paper, we deliver three key contributions:

\begin{enumerate}
	\item \textbf{A broadly-applicable, agentic generation framework.} We open-source a "World-Author–Critic–Validator" pipeline that can embed any deterministically-verifiable reasoning graph—not only arithmetic ListOps but also symbolic, abductive, inductive, or defeasible logic (see Appendix \ref{sec:extending_VLO}).
	\item \textbf{Verbose ListOps (VLO): a first instantiation of the framework.} Using this pipeline we build VLO, a benchmark that narrativises nested ListOps computations.  VLO is the first long-context dataset that (i) withholds all intermediate results, (ii) controls context length and reasoning depth orthogonally, and (iii) ships exact intermediate-step labels, enabling systematic failure-mode analysis.
	\item \textbf{Orthogonal control knobs for context and reasoning complexity.} Tree depth, operator arity, narrative length, and distraction density are exposed as independent hyper-parameters, allowing researchers to stress-test scaling behaviour along a single axis while holding all others fixed—something prior benchmark suites cannot do.
	\item \textbf{A diagnostic evaluation toolkit and empirical study.} We release scripts that check correctness at every anchor step, plot error-propagation curves, and compare prompting regimes.  Applying this toolkit to ten prominent LLMs uncovers a consistent blind spot: models that excel at raw ListOps or factual recall often diverge at the very first hidden step when the same logic is wrapped in narrative camouflage.
\end{enumerate}

Together, the framework plus toolkit turn VLO from a one-off dataset into a reusable test-bed for any future long-context reasoning architecture.

Our experiments show state-of-the-art LLMs, despite solving ListOps with ease, on VLO, suffer a \textbf{severe ($\approx$50\%+) drop in accuracy at just $\approx$10k-token contexts}. The result exposes an under-tested limitation: LLMs struggle to synthesize multi-step conclusions amongst narrative distraction. VLO offers a rigorous, open-source tool for diagnosing (and improving) this foundational capability, seeking to move discussion beyond simply enlarging context windows.

\section{Related Work}
The expansion of LLM context windows has catalyzed research into their performance on extensive textual inputs, leading to a diverse landscape of evaluation benchmarks.

\setlength{\tabcolsep}{2pt}
\renewcommand{\arraystretch}{0.9}
\begin{table}[htbp]
	\centering
	\scriptsize
	\resizebox{\textwidth}{!}{%
		\begin{tabular}{%
				L{1.8cm}   
				L{2cm}   
				C{1.2cm}   
				L{1.9cm}   
				C{1.3cm}   
				C{1.1cm}   
				L{1.6cm}   
				C{1cm}     
				L{1.2cm}   
			}
			\toprule
			\makecell[c]{\textbf{Benchmark}}                     &
			\makecell[c]{\textbf{Primary}                          \\\textbf{Reasoning}\\\textbf{Task}} &
			\makecell[c]{\textbf{Nested}                           \\\textbf{Reasoning}\\\textbf{Chains}} &
			\makecell[c]{\textbf{Distraction}}                   &
			\makecell[c]{\textbf{Tunable}                          \\\textbf{Reasoning}\\\textbf{Difficulty}} &
			\makecell[c]{\textbf{Scalable}                         \\\textbf{Context}} &
			\makecell[c]{\textbf{Deterministic}                    \\\textbf{Answer}} &
			\makecell[c]{\textbf{Realistic}                        \\\textbf{Tasks}} &
			\makecell[c]{\textbf{Generation}                       \\\textbf{Method}} \\
			\midrule
			Verbose ListOps (Ours)                               &
			Algorithmic (ListOps) embedded in narrative          &
			\yesmark                                             &
			Coherent, semantically relevant narrative            &
			\yesmark                                             &
			\yesmark                                             &
			Mathematically deterministic                         &
			\yesmark                                             &
			Agentic                                                \\
			\addlinespace
			LongReason (\citeyear{ling2025longreason})           &
			General QA (Comprehension, Inference, Maths)         &
			\nomark                                              &
			Irrelevant passages embedded around relevant context &
			\somewhatmark                                        &
			\yesmark                                             &
			Multiple-choice (via reasoning)                      &
			\yesmark                                             &
			Context expansion                                      \\
			\addlinespace
			Needle-in-Haystack Type                              &
			Factual Recall                                       &
			\nomark                                              &
			Vast irrelevant corpus                               &
			\nomark                                              &
			\yesmark                                             &
			Exact match                                          &
			\nomark                                              &
			Target insertion                                       \\
			\addlinespace
			\makecell[l]{Other Synthetic                           \\(e.g. RULER \\ \citeyear{hsieh2024ruler_arxiv}.)} &
			\makecell[l]{Code/Instruction                          \\Following} &
			\nomark                                              &
			Structured, less narrative                           &
			\nomark                                              &
			\yesmark                                             &
			Task-specific                                        &
			\nomark                                              &
			Template Based                                         \\
			\addlinespace
			Human Annotated Long-Context QA                      &
			General QA                                           &
			\nomark                                              &
			Natural document structure                           &
			\nomark                                              &
			\nomark                                              &
			Human-judged                                         &
			\yesmark                                             &
			Human annotation                                       \\
			\bottomrule
		\end{tabular}
	}

	\vspace{8pt}
	\caption{Comparison of Verbose ListOps with other long-context benchmarks. Using an agentic generation process, Verbose ListOps offers both controllable context lengths and reasoning difficulty.}
	\label{tab:benchmark_comparison_intro}
\end{table}

\subsection{Long-Context Benchmarks}
Early long-context evaluations typically adapted standard NLP benchmarks \citep{shaham-etal-2023-zeroscrolls, an2023l, bai-etal-2024-longbench2}. These early benchmarks often had contexts shorter than the maximum capabilities of current LLMs \citep{gemini2024unlocking, openai2025gpt41_api, anthropic2025claude37sonnet_blog_actual} and inadequately differentiated complex reasoning tasks from simpler tasks based on factual recall tasks or failed to analyze the impact of distractors sufficiently \citep{muhlgay2023factor}.

Recent synthetic benchmarks have addressed some of these limitations by allowing greater control over context length. The "needle-in-a-haystack" (NIAH) paradigm, exemplified by NeedleBench \citep{li_mo2024needlebench}, specifically evaluates an LLM's recall capability within extensive irrelevant contexts, focusing on fact extraction rather than complex multi-step reasoning. Other synthetic benchmarks, such as RULER \citep{hsieh2024ruler_arxiv}, test specific reasoning capabilities like variable tracking and multi-hop information extraction. However, their artificial scenarios may not adequately replicate the challenges posed by naturalistic narratives with coherent distractors \citep{haller2024pecc}.

Benchmarks such as InfiniteBench \citep{zhang2024infinitebench_acl} or those utilizing extensive document curation like LooGLE \citep{li-etal-2024-loogle} often rely heavily on human annotation or semi-automated methods. While realistic, these approaches are labor-intensive and limit scalability and precise control over task variables such as context length and complexity, underscoring the need for standardized, synthetic evaluation frameworks \citep{valmeekam2023planbench, muhlgay2023factor}.

LongReason \citep{ling2025longreason} significantly advances synthetic benchmarks by offering variety—reading comprehension, logical inference, and mathematical reasoning tasks are expanded into longer, distractor-rich texts. It challenges models to aggregate and reason over scattered information. In contrast, the VLO benchmark presents a distinct challenge by embedding nested algorithmic ListOps problems within coherent narratives, uniquely emphasizing internal computation and state management. This approach differs fundamentally from LongReason, which centers on aggregating explicitly presented clues rather than tracking values that must be computed from prior steps. 

\subsection{The Emergence of Reasoning alongside Long Context}
As strong large-context capabilities became the norm, demand grew for more complex reasoning within those extended windows, fueling the development of "reasoning" models and benchmarks. Surveys of long-context benchmarks show LLMs recall facts across tens of thousands of tokens but falter at multi-step inference over extended inputs. \cite{liu-etal-2024-modeling} find ultra-long transformers (100k+ tokens) struggle to integrate dispersed information as context grows. \cite{hsieh2024ruler_arxiv} report similar declines on RULER tasks, highlighting that wider windows alone don't guarantee accurate intermediate-state tracking. \cite{shi2023large} term this "attention dilution," where relevant facts become difficult to aggregate in long contexts. Consequently, current research now emphasizes explicit planning, modular computation, or memory mechanisms.

\paragraph{Prompting‐Based Reasoning Techniques.}
The first widely adopted method to elicit stepwise reasoning from LLMs was Chain-of-Thought (CoT) prompting, instructing models to think step-by-step or via in-context examples, enhancing arithmetic and logic performance without fine-tuning \citep{wei2022chainofthought_neurips}. However, CoT becomes brittle with deep nesting or backtracking, causing cascading errors. \cite{yao2023treeofthoughts_neurips} propose Tree-of-Thoughts, which explores multiple reasoning branches in parallel, uses verification to prune incorrect paths, and allows backtracking. This search-based method helps models recover from missteps and leverage dispersed information more effectively.

\paragraph{Architectures for Explicit Reasoning.}
Beyond prompt engineering, recent architectures directly embed planning and state-tracking. \emph{Large Concept Models} \cite{lcm_team2024concept_arxiv} process higher-order semantic units rather than text tokens, tracking each reasoning step as a 'concept'. Similarly, \cite{mondal2024pfc_planning} introduce a prefrontal cortex--inspired Modular Agentic Planner (MAP), where a central LLM orchestrates modules proposing, evaluating, and refining subgoals via a shared scratchpad to track intermediate states. Outperforming single-pass transformers on classical benchmarks (e.g., Tower of Hanoi), MAP highlights benefits to decoupling planning from execution.

\paragraph{Memory‐Augmented and Retrieval‐Augmented Reasoning.}
Architectures with explicit memory modules show promise over long‐context reasoning. Recurrent Memory Transformers (RMTs) carry summaries of past hidden states across segments to avoid reprocessing. In BABI‐Long evaluations, RMTs outperform standard long‐attention models by integrating clues over 128k tokens \citep{wang2024loong_realworld}. Retrieval‐augmented generation (RAG) complements these: by fetching relevant passages, LLMs ground inferences (e.g., \cite{an2023l}). However, \cite{hengle2025mlmextended} show RAG pipelines—despite high retrieval accuracy—often fail when chained reasoning steps are needed in extended contexts. Likewise, the LaRA benchmark (2025) finds long‐context LLMs outperform RAG on multi‐step arithmetic problems as document length scales \citep{cai2025lara}.

As VLO exposes every intermediate state, it is also a natural playground for Process-Reward Models (PRMs) that score reasoning chains rather than final outputs. We expect the dataset to serve both as a training signal and as a forensic tool for such chain-inspection methods.

\section{Verbose ListOps (VLO): Specification and Construction}
\label{sec:VLO-spec}

This section formalises the Verbose ListOps (VLO) task, details its programmatic construction pipeline, and presents the controllable parameters used to stress‑test long‑context language models.

\subsection{Formal Task Definition}
\label{subsec:task-def}

Given a narrative $X \in \mathbb{T}_{10\mathrm{k}\text{-token}}$ and an implicit ListOps abstract syntax tree (AST) $T$ with leaves (atomic integers) and internal nodes (operators), the original ListOps task \citep{nangia2018listops_sereTOD} requires models to evaluate $T$ by performing its specified operations (e.g., \texttt{SUM}s, \texttt{MIN}imums, \texttt{MEDIAN}s) on numerical inputs to yield a deterministic result. For example, if

\[
	T = \texttt{MAX}(\texttt{SUM}(2,1),\,4),
\]

then one computes $\texttt{SUM}(2,1)=3$ and subsequently $\texttt{MAX}(3,4)=4$.

VLO embeds each nested ListOps operation in a narrative, describing each node in $T$ via post-order traversal. Placeholders $\{a_i\}$ (which we call 'narrative anchors') refer to intermediate values without explicitely stating the results.

Formally, the VLO task is:

\[
	f\colon X \longrightarrow v(T),
\]

…where v(T) denotes the scalar value at the root of the abstract-syntax tree (i.e. the final numerical result). Success is $f(X)$ matching the ground-truth integer.

\paragraph{Running example}
Consider the above ListOps problem again. Intermediate values appear only as anchors ${a_i}$. The first operation is

\[
	\texttt{SUM}(2,1) \;\to\; a_1,\quad a_1 = 3
\]

In VLO, this operation would be narrativized as the following:

\begin{quote}
	"In the bustling spaceport of Xylos, Chief Engineer Anya cataloged incoming parts. Her logs showed \textbf{one} crate of cryo‐cells arrived on the morning freighter, and later, \textbf{two} additional crates were offloaded from a fast courier. She processed these figures, which she labeled as \textit{'Daily Cell Intake'} (anchor $a_1$) in preperation for her presentation at Xylos' annual Meteor Shower Festival.
\end{quote}
\begin{quote}
	\textit{Notice \textit{'Daily Cell Intake'} represents $a_1$ (the first operation's result), done so the number `3' (result of the \texttt{SUM}) never appears.}
\end{quote}

The second operation is

\[
	\texttt{MAX}(a_1,4) \;\to\; a_2,\quad a_2 = 4
\]

…yielding the final result $v(T)$. The narrative then compares \textit{'Daily Cell Intake'} ($a_1$) with four, names the larger as anchor $a_2$, and omits that $a_1=3$. This preserves the no‐numeric‐leakage constraint and ensures the model's output equals $v(T)$.

\subsection{Design Constraints}
\label{subsec:design-constraints}
VLO enforces three orthogonal constraints that distinguish it from prior long‑context benchmarks:
\begin{description}
	\item[C1 No numeric leakage.] Intermediate results are \emph{never stated}; they may only be referenced via anchors (e.g., ``the daily cell intake''). Explicit numerals for non‑leaf nodes are prohibited.
	\item[C2 Narrative camouflage.] LLMs weave operator descriptions into a coherent story that contains LLM-generated, semantically related computation‑irrelevant content (distractors), mirroring real‑world documents where crucial information is buried among related context.
	\item[C3 Parametric control.] ListOps complexity (operator set, tree depth $d$, branching factor $b$) is adjustable, enabling ablations; increasing $d$ or $b$ adds anchors and narrative density.
\end{description}

\subsection{Generation Pipeline}
\label{subsec:generation}
\begin{figure}[ht]
	\centering
	\includegraphics[width=1\textwidth]{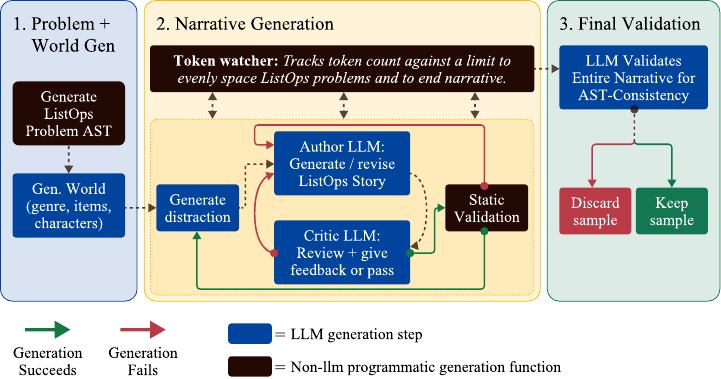}
	\caption{The Verbose ListOps (VLO) agentic generation and validation pipeline. The process begins with AST sampling and world generation, then iteratively constructs narrative segments using Author-Critic LLM pairs with programmatic validation, incorporates distraction content, and concludes with holistic validation to ensure sample validity and adherence to constraints C1-C3.}
	\label{fig:pipeline}
\end{figure}

Figure~\ref{fig:pipeline} outlines how the VLO construction pipeline ensures adherence to constraints C1–C3 through the following five-stage process (implementation details and prompts are provided in Appendices~\ref{app:generation_pipeline} and \ref{app:prompts}):

\begin{enumerate}
	\item \textbf{AST Sampling.} Sample a ListOps AST $T$ of depth $d$ and branching factor $b$, and compute its root value $v(T)$ by deterministic post‐order evaluation.
	\item \textbf{World Generation.} Generate fictional world metadata (genre, setting, characters, and primary objects) using an LLM to establish thematic coherence for narrative embedding (constraint C2). This creates the contextual foundation for embedding ListOps operations into a cohesive story. \emph{Narrative anchors}—conceptual names like "Daily Cell Intake"—are generated to reference intermediate results without revealing numerical values (constraint C1). Detailed parameters are in Appendix~\ref{app:generation_hyperparams}.
	\item \textbf{Distraction Generation.} Generate semantically relevant but computation-irrelevant content between ListOps operations. This expands context length while forcing models to filter computational information from extraneous narrative detail, using the same world context to maintain coherence without introducing interfering numerical information that can interfere with the underlying reasoning task. The \emph{token watcher} monitors narrative length throughout generation, dynamically allocating the remaining token budget after ListOps narratives.
	\item \textbf{Operator Narrative Construction.} For each ListOps node, build a narrative segment via:
	      \begin{itemize}
		      \item[(a)] \textbf{Author Agent.} The Author LLM drafts a segment for the current operator $\omega \in \mathcal{O}$ and its child anchors $a_i$ (if applicable), weaving it into the story (C2) while avoiding numeric mention of the current result (C1) and respecting parameters (C3).
		      \item[(b)] \textbf{Critic Agent.} The Critic LLM reviews the draft, checking:
		            \begin{itemize}
			            \item[\textbullet] No numeric digits or words for the current operation's result appear (enforces C1).
			            \item[\textbullet] Correct anchor use (for child inputs of prior operations) and atomic inputs.
			            \item[\textbullet] Narrative coherence (ensures C2).
		            \end{itemize}
		            It proposes edits if it finds violations, and the Author revises until all checks pass.
		      \item[(c)] \textbf{Programmatic Static Validation.} A script scans the finalized segment to verify:
		            \begin{itemize}
			            \item[\textbullet] Absence of any digit or word numbers for the current operation's result.
			            \item[\textbullet] Presence of anchor tokens (if expected) or numerical atomic inputs.
			            \item[\textbullet] No unexpected placeholders or formatting.
		            \end{itemize}
		            Any failure triggers a re‑generation of that segment.
	      \end{itemize}
	\item \textbf{Comprehensive LLM Validation:} A larger Validator LLM, aided by extensive prompting and few-shot examples, performs step-by-step evaluation against the original AST $T$ to detect numeric leaks, verify correct representation of operations and inputs, check for missing/duplicate operators, ensure coherence (C2), and confirm ground truth value $v(T)$.
	      Violations identified by the Validator LLM cause the sample to be discarded and resampled.
\end{enumerate}

\subsection{Controllable Parameters}
\label{subsec:params}
VLO's generation pipeline allows fine‑grained difficulty control via \texttt{Config}. Key parameters include:
\begin{itemize}
	\item \textbf{Narrative length:} The target length of a sample.
	\item \textbf{Tree complexity:} Parameters such as tree depth $d$ (\texttt{MAX\_OPS}) and operator branching factor $b$ (arity, controlled by \texttt{MIN\_ARITY} and \texttt{MAX\_BRANCH}) are tunable. For the specific settings used to generate the dataset for this paper, please refer to Section~\ref{subsec:exp-setup} and Appendix~\ref{app:generation_hyperparams}. Larger $d$ or $b$ generally increases the number of anchors and thus the narrative density.
\end{itemize}

\subsection{Evaluation Protocol}
\label{subsec:evaluation}
Models receive the full narrative $X$ (with anchors $\{a_i\}$) and must output an integer. We report exact‑match accuracy with 95\% Wilson confidence intervals, as in recent long‑context benchmarks ~\citep{an2023l, zhang2024infinitebench_acl}. \texttt{evaluator.py} conducts the evaluation (and \texttt{equation\_llm\_evaluator.py} for standard ListOps). Chain‑of‑thought or external tool use is \emph{prohibited} to isolate internal computation. All code, generation logs, and datasets are open‑source.

\subsection{Experimental Setup}
\label{subsec:exp-setup}
\begin{itemize}
	\item \textbf{Models.} We test Gemini 2.5 Pro and Flash, OpenAI o4-Mini and GPT-4.1, Grok 3-Mini, Claude 3.7 Sonnet, DeepSeek R1 and V3, Qwen-3 235B, and Llama-4 Maverick.
	\item \textbf{VLO instances.} 1,000 VLO-10k samples were generated at a cost of $\approx\$1,500$~USD. The embedded ListOps problems within these instances used the following parameters: max 8 operations per AST, max branching factor of 8, min operator arity of 4, and atomic integer values from range [1, 30]. Full hyperparameter details are available in Appendix~\ref{app:generation_hyperparams}.
	\item \textbf{Standard ListOps baseline.} For each VLO sample, we construct the bare ListOps expression, providing an algorithmic baseline of identical difficulty without narrative distractors.
	\item \textbf{Evaluation inference config.} All models had $temperature = 0.01$ and $top\_p = 0.95$ and were accessed via OpenRouter. Total evaluation incurred a cost of $\approx\$500$~USD.
\end{itemize}

\subsection{Results}
\label{subsec:results}
The table below reports exact-match accuracy on both Verbose ListOps and its corresponding bare ListOps expressions alongside performance on the OpenAI MRCR NIAH benchmark. Whilst nearly every 'thinking' model achieves perfect or near-perfect accuracy on bare ListOps, performance collapses under Verbose ListOps \emph{at just 10-k tokens}.

\begin{table}[htbp]
	\centering
	\scriptsize
	{
		\setlength{\arrayrulewidth}{0.01pt}
		\setlength{\tabcolsep}{1em}
		\begin{tabular}{l|ccc|cc}
			\toprule
			\textbf{Model}            & \textbf{ListOps} & \textbf{VLO-10k} & \textbf{OAI MRCR-8k 8-Pin}\footnotemark & \textbf{95\% CI (ListOps)} & \textbf{95\% CI (VLO)} \\
			\midrule
			\multicolumn{6}{c}{\itshape Closed-source models}                                                                                                               \\ [2pt]
			Gemini 2.5 Pro            & 100.0            & 55.3             & 86.0                                    & 99.6–100.0                 & 52.2–58.4              \\
			Gemini 2.5 Flash Thinking & 100.0            & 49.1             & 63.3                                    & 99.6–100.0                 & 46.0–52.2              \\
			o4 Mini High              & 100.0            & 48.1             & 63.6                                    & 99.6–100.0                 & 45.0–51.2              \\
			Grok 3 Mini High          & 81.7             & 47.2             & 51.6                                    & 79.2–84.0                  & 44.1–50.3              \\
			Claude 3.7 Sonnet High    & 58.4             & 40.0             & N/A                                     & 55.3–61.4                  & 37.0–43.1              \\
			GPT-4.1                   & 38.6             & 32.2             & 29.6                                    & 35.6–41.7                  & 29.3–35.2              \\
			\midrule
			\multicolumn{6}{c}{\itshape Open-source models}                                                                                                                 \\ [2pt]
			DeepSeek R1               & 98.4             & 41.2             & N/A                                     & 97.4–99.1                  & 38.2–44.3              \\
			Qwen-3 235B               & 53.7             & 41.5             & N/A                                     & 50.6–56.8                  & 38.4–44.6              \\
			Llama 4 Maverick          & 47.2             & 32.0             & N/A                                     & 44.1–50.3                  & 29.1–35.0              \\
			DeepSeek V3 0324          & 93.7             & 25.1             & N/A                                     & 92.0–95.1                  & 22.4–28.0              \\
			\bottomrule
		\end{tabular}
	}
	\vspace{4pt}
	\caption{Accuracy on VLO-10k versus its corresponding bare ListOps expressions, based on evaluation of 1,000 samples per model. \emph{Note DeepSeek V3's drop from 93.7\% to 25.1\%.}}
	\label{tab:model_performance_results}
\end{table}

\footnotetext{OpenAI's Multi-Reference Context Recall (MRCR) benchmark: 8-passage subset measuring pure retrieval accuracy under 8k tokens without multi-step reasoning. Provides baseline for comparing retrieval vs. VLO's narrative-embedded reasoning. Results from ContextArena \citep{ContextArena2025}.}

\subsection{Analysis and Limitations}
\label{subsec:discussion}

\paragraph{Narrative Camouflage, 'Cognitive Control', and Architectural Divergence}
VLO challenges models to find, solve, track, and synthesize nested reasoning problems amongst narrative camouflage to answer a deterministic question, creating a information processing burden that exposes fundamental architectural differences between models. The results in Table~\ref{tab:model_performance_results} reveal two critical takeaways. First, models with explicit reasoning scaffolds (e.g., Gemini 2.5, o4-Mini) perform better on VLO than those without. Second, the stark contrast between DeepSeek-V3's near-perfect score on raw ListOps and its collapse on VLO demonstrates while scratchpad-like mechanisms are not required for nested reasoning, they are crucial for filtering noise and protecting the reasoning process from distractors.

This architectural divergence explains the performance trends. DeepSeek-V3's design, while highly efficient, creates specific vulnerabilities to VLO's challenge. Its architecture is optimized for performance on standard benchmarks through two key strategies that, we hypothesize, compromise its task resilience on VLO. First, its \textit{auxiliary-loss-free load balancing} for its Mixture-of-Experts (MoE) architecture encourages over-specialisation, where the gating network routes inputs to a "narrative" expert that fails to pattern-match the embedded logic. Second, its use of \textit{Multi-Token Prediction (MTP)} encourages the model to "pre-plan" its output based on narrative flow, reinforcing the very heuristic processing that VLO penalizes \citep{deepseek2025deepseekv3}.

Crucially, DeepSeek-V3's technical report clarifies that its math and coding reasoning capabilities stem not from emergence but from \emph{distillation from DeepSeek-R1}, which has long-Chain-of-Thought capabilities. This process "notably improves its reasoning performance" by adopting R1's "verification and reflection patterns" into V3 \citep{deepseek2025deepseekv3}. Therefore, V3's reasoning is a specialized, distilled skill rather than an inherent, flexible process—explaining its brittleness: these patterns excel on structured tasks like raw ListOps but falter under VLO's narrative camouflage, which demands a more robust, first-principles reasoning scaffold.

In contrast, Gemini's architecture, founded on a highly efficient sparse Mixture-of-Experts (MoE) architecture \citep{gemini2024unlocking}, appears to incorporate more robust mechanisms for maintaining task focus amidst distraction. An analysis of Gemini 2.5's 'thinking' on VLO suggests its MoE implementation may avoid this hyper-specialization trap. When processing a VLO problem, it behaves as if it first decomposes the task, managing the narrative, operands, and operators as distinct variables. Its MoE routing is then conditioned on this sub-task, and behaves in a way that seems to dynamically shift from activating semantic experts for the narrative to logic-and-reasoning experts for the calculation, in what can be described as dynamic context-aware routing. This ability to parse and apply a formal, rule-based system provided entirely in-context is the same fundamental capability demonstrated in the Gemini 1.5 \emph{Technical Report}, where the model learns to translate Kalamang—a language with <200 speakers—by processing an in-prompt 500-page grammar book and dictionary \citep{gemini2024unlocking}.

Viewing this delta as a proxy for maintaining appropriate attention over long contexts, this architectural split can explain the shrinking performance delta between MRCR and VLO for less capable 'thinking' models. Less capable models have uniformly weak heuristic processing—not sophisticated enough to be hijacked by narrative—and too weak algorithmic execution, yielding two low, closely clustered scores. Conversely, Gemini 2.5's architecture explains its large delta: its heuristic system excels at MRCR but generates interference that its internal reasoning scaffold struggles to overcome on VLO, a known LLM issue \citep{shi2023large, vishwanath2025meddistractqa}. This aligns with findings that LLMs struggle to shift from intuitive, pattern-matching reasoning to deliberate, step-by-step processing for unfamiliar or complex structures \citep{mirzadeh2024gsm_symbolic}. Thus, the performance gap between recall and narrative reasoning powerfully proxies this architectural conflict.

\paragraph{Limitations and Future Work}
Using a fixed $\approx$10k-token narrative exploits the strong recall abilities of SOTA LLMs at this context size \citep{ContextArena2025}, allowing VLO to isolate reasoning deficits from failures in factual recall. The degradation observed (Table 2) at a context length where recall is robust underscores VLO's focus on narrative-embedded computational challenges. Future studies should vary context length more broadly to explore these dynamics.

Further considerations and avenues for future research include:

\begin{itemize}
	\item \textbf{Generator Model Bias:} VLO-10k was generated with Gemini 2.5.\footnote{During development we found only Gemini 2.5 and GPT-4.5 could reliably generate VLO, with GPT-4.5 pricing being cost-prohibitive (\$75USD/M input tokens, \$150USD/M output tokens \citep{openai_pricing_2025}).} This introduces potential generator model bias, where evaluated models from the same family may exhibit inflated performance due to stylistic or implicit knowledge alignment. Future work should investigate this and explore mitigation strategies, such as employing a more diverse set of future generator models or incorporating bias-neutralization techniques \citep{Yuan2025Silencer}.
	\item \textbf{Chain-of-Thought (CoT) Prohibition:} This evaluation prohibits external CoT prompting to isolate core computational reasoning as today's 'thinking' models have internal CoT-like functions. Given DeepSeek v3's notable degradation on VLO, a future evaluation allowing external CoT or advanced prompting (e.g., Tree-of-Thoughts \cite{yao2023treeofthoughts_neurips}) could clarify if explicit scratchpad prompting restores reasoning or reveals inherent architectural limits.
	\item \textbf{Depth of Failure Mode Analysis:} While Table~2 compellingly demonstrates significant performance drops, the current analysis remains primarily quantitative. A granular error analysis would help explain \textit{why} models fail on VLO tasks—whether due to narrative parsing errors, incorrect ListOps computations, or challenges in tracking internal states amid distractors. Investigating error cascading or differences by operator type, tree complexity, or narrative structure could inform the development of more robust reasoning architectures.
	\item \textbf{Scope and Generalizability of ListOps:} Currently, VLO uses ListOps to evaluate algorithmic execution and state-tracking within narratives. This initial focus does not capture the full range of real‐world narrative reasoning, which involves ambiguity, implicit knowledge, and logical frameworks like abductive, inductive \citep{sheng-etal-2025-evaluating, bowen2024comprehensiveinductive}, or defeasible reasoning \citep{ren2024symtex, Leidinger2024LLMsNonmonotonic}. Fortunately, the VLO generation pipeline is highly extensible, and can support symbolic non-numeric problems, enabling future variants to test reasoning such as abducting causes, inducing general rules, or handling defeasible updates—providing a more comprehensive LLM evaluation. See Appendix~\ref{sec:extending_VLO} for details on extending VLO to these other reasoning types.

\end{itemize}

\paragraph{Key takeaway.}
Sustaining multi-step computation amidst noise requires architectures that explicitly model reasoning steps. Parameter scale or context window size alone cannot bridge this weakness.

\section{Discussion}
VLO was initially developed to benchmark automated predictive signal extraction from distributed narratives (e.g., assessing prospect 'consequence awareness' from sales communications). Results here highlight a critical gap in automating sophisticated analytics: failures observed in VLO mirror real-world issues, such how a lawyer might lose track of interacting clauses or salespeople misinterpreting intent due to narrative complexity. State-of-the-art LLMs excel at fact recall but falter when reasoning with extracted facts, underscoring that improved context length alone is insufficient. VLO thus provides an essential testbed for emerging architectures capable of addressing these reasoning challenges, including those with explicit planning (e.g., PFC-inspired systems \citep{mondal2024pfc_planning}) and advanced prompting strategies like \cite{yao2023treeofthoughts_neurips}'s Tree-of-Thoughts.

Concept-oriented Large Concept Models (LCMs),\footnote{'LCMs' here \textbf{do not} refer to Latent Consistency Models \citep{luo2023latent} used for image synthesis.}  which operate on higher-level semantic units rather than individual tokens \citep{lcm_team2024concept_arxiv} offer a promising direction. For VLO, an LCM treats each reasoning step or conceptual reference to an intermediate result (our "narrative anchors") as distinct conceptual units. This aligns more closely with how humans might track such information and could make the model more resistant to narrative distractors when trying to maintain the integrity of a numerical value associated with a concept. In performing autoregressive prediction in a concept embedding space, rather than a token space, LCMs could facilitate more stable tracking of multi-step computations. Such concept embeddings can be a robust and less diffuse carrier of the numerical value of an intermediate result than distributed token activations.

Finally, VLO's programmatic framework shows promise in how it agentically embeds deterministic tasks into narratives to force internal computation. Such a method can be extended to any domain requiring implicit criteria encoding (done here via narrative anchors) to create more realistic and challenging synthetic datasets, e.g., LLM-as-Judge, Multi-Hop QA, Code-Based Evaluation, and Long-Context QA/Summarization evaluations. VLO thus not only exposes current limitations but also offers a methodology for developing next-generation LLMs with deeper reasoning.

\subsection*{Broader Impacts}
VLO pushes for Large Language Models that can effectively process long, complex narratives. Such advances are stepping stones for LLMs to unearth predictive signals hidden in unstructured text. The upsides are clear: sharper analytical tools for science, more insightful legal review, or more astute financial risk assessment from textual data. Our effort to pinpoint and fix current reasoning flaws is about building more dependable Artificial Intelligence for these demanding roles.

However, creating LLMs skilled at deciphering narratives raises significant dual-use concerns. The capacity to extract predictive signals from text, while beneficial, could be repurposed for high-stakes social monitoring. This creates a direct risk of what is often termed predictive policing, where models could generate pre-emptive, and potentially biased, judgments about individuals based on textual data \citep{eucpn2022predictive_policing}. If an AI claims to "predict" behavior from text, the door opens to profiling and discrimination, creating a climate of surveillance that could chill free expression—a core concern in the literature on surveillance capitalism \citep{zuboff2019surveillance_capitalism}.

Further, relying on LLMs for prediction is risky when their interal logic is opaque: unexpected failures increase in likelihood, with these errors being hard to detect or correct \citep{rane2024blackbox_xai}. Hence, high stakes 'automated signal extraction' gone wrong can deeply affect lives. AI-driven economic shifts likewise redefine professional roles through task-based automation rather than simple job loss, polarizing the labor market: routine analytical tasks may be automated while experts who manage, interpret, and validate these systems become more in demand \citep{autor2020work_of_future}.

VLO directly contributes to the development of more scrutable and robust AI. Opaque reasoning can lead to biased outcomes or unexplained failures in critical domains—problems that require tools to pinpoint and falsify specific reasoning paths. The risks of opaque reasoning—from biased predictions to unexplainable failures in safety-critical domains—cannot be mitigated without tools that allow for the falsification of specific reasoning pathways. VLO offers such a tool: a controlled, deterministic envrionment where a model's failures reveal precise breakdowns in its computation rather than just an accuracy drop. By isolating narrative-embedded reasoning, it delivers a clear diagnostic signal for architectural and algorithmic improvements. This supports the community's shift from merely scaling capabilities to building trustworthy systems \citep{stanford_hai_2025_ai_index}. Responsible AI development demands models that are not only powerful but also interpretable—and VLO helps us understand exactly where and why they fail.

\bibliographystyle{plainnat}
\bibliography{\jobname}

\begin{thebibliography}{41}
\providecommand{\natexlab}[1]{#1}
\providecommand{\url}[1]{\texttt{#1}}
\expandafter\ifx\csname urlstyle\endcsname\relax
  \providecommand{\doi}[1]{doi: #1}\else
  \providecommand{\doi}{doi: \begingroup \urlstyle{rm}\Url}\fi

\bibitem[An et~al.(2023)An, Gong, Zhong, Zhao, Li, Zhang, Kong, and Qiu]{an2023l}
Chenxin An, Shansan Gong, Ming Zhong, Xingjian Zhao, Mukai Li, Jun Zhang, Lingpeng Kong, and Xipeng Qiu.
\newblock L-eval: Instituting standardized evaluation for long context language models.
\newblock \emph{arXiv preprint arXiv:2307.11088}, 2023.

\bibitem[{Anthropic}(2025)]{anthropic2025claude37sonnet_blog_actual}
{Anthropic}.
\newblock {Introducing the Claude 3.7 Model Family}.
\newblock \url{https://www.anthropic.com/news/claude-3-7}, 2025.
\newblock Accessed: 2025-05-15.

\bibitem[Autor et~al.(2020)Autor, Mindell, and Reynolds]{autor2020work_of_future}
David Autor, David~A. Mindell, and Elisabeth~B. Reynolds.
\newblock {The Work of the Future: Building Better Jobs in an Age of Intelligent Machines}.
\newblock Technical report, {MIT Task Force on the Work of the Future}, Cambridge, MA, 2020.

\bibitem[Bai et~al.(2024)Bai, Tu, Zhang, Peng, Wang, Lv, Cao, Xu, Hou, Dong, Tang, and Li]{bai-etal-2024-longbench2}
Yushi Bai, Shangqing Tu, Jiajie Zhang, Hao Peng, Xiaozhi Wang, Xin Lv, Shulin Cao, Jiazheng Xu, Lei Hou, Yuxiao Dong, Jie Tang, and Juanzi Li.
\newblock {LongBench v2: Towards Deeper Understanding and Reasoning on Realistic Long-context Multitasks}.
\newblock \emph{{arXiv} preprint arXiv:2412.15204}, 2024.

\bibitem[Bowen et~al.(2024)Bowen, S{\ae}tre, and Miyao]{bowen2024comprehensiveinductive}
Chen Bowen, Rune S{\ae}tre, and Yusuke Miyao.
\newblock A comprehensive evaluation of inductive reasoning capabilities and problem solving in large language models.
\newblock In \emph{Findings of the Association for Computational Linguistics: EACL 2024}, pages 323--339, St. Julian's, Malta, March 2024. Association for Computational Linguistics.
\newblock URL \url{https://aclanthology.org/2024.findings-eacl.22}.

\bibitem[DeepSeek-AI et~al.(2025)DeepSeek-AI, Liu, Feng, Xue, Wang, Wu, Lu, Zhao, Deng, Zhang, Ruan, Dai, Guo, Yang, Chen, Ji, Li, Lin, Dai, Luo, Hao, Chen, Li, Zhang, Bao, Xu, Wang, Zhang, Ding, Xin, Gao, Li, Qu, Cai, Liang, Guo, Ni, Li, Wang, Chen, Chen, Yuan, Qiu, Li, Song, Dong, Hu, Gao, Guan, Huang, Yu, Wang, Zhang, Xu, Xia, Zhao, Wang, Zhang, Li, Wang, Zhang, Zhang, Tang, Li, Tian, Huang, Wang, Zhang, Wang, Zhu, Chen, Du, Chen, Jin, Ge, Zhang, Pan, Wang, Xu, Zhang, Chen, Li, Lu, Zhou, Chen, Wu, Ye, (Ye, Ma, Wang, Zhou, Yu, Zhou, Pan, Wang, Yun, Pei, Sun, Xiao, Zeng, Zhao, An, Liu, Liang, Gao, Yu, Zhang, Li, Jin, Wang, Bi, Liu, Wang, Shen, Chen, Zhang, Chen, Nie, Sun, Wang, Cheng, Liu, Xie, Liu, Yu, Song, Shan, Zhou, Yang, Li, Su, Lin, Li, Wang, Wei, Zhu, Zhang, Xu, (duplicate removed), Huang, Li, Zhao, Sun, Li, Wang, Yu, Zheng, Zhang, Shi, Xiong, He, Tang, Piao, Wang, Tan, Ma, Liu, Guo, Wu, Ou, Zhu, Wang, Gong, Zou, He, Zha, Xiong, Ma, Yan, Luo, You, Liu, Zhou, Wu, Ren, Ren, Sha, Fu, Xu, Huang, Zhang, Xie, Zhang, Hao, Gou, Ma, Yan, Shao, Xu, Wu, Zhang, Li, Gu, Zhu, Liu, Li, Xie, Song, Gao, and Pan]{deepseek2025deepseekv3}
DeepSeek-AI, Aixin Liu, Bei Feng, Bing Xue, Bingxuan Wang, Bochao Wu, Chengda Lu, Chenggang Zhao, Chengqi Deng, Chenyu Zhang, Chong Ruan, Damai Dai, Daya Guo, Dejian Yang, Deli Chen, Dongjie Ji, Erhang Li, Fangyun Lin, Fucong Dai, Fuli Luo, Guangbo Hao, Guanting Chen, Guowei Li, H.~Zhang, Han Bao, Hanwei Xu, Haocheng Wang, Haowei Zhang, Honghui Ding, Huajian Xin, Huazuo Gao, Hui Li, Hui Qu, J.~L. Cai, Jian Liang, Jianzhong Guo, Jiaqi Ni, Jiashi Li, Jiawei Wang, Jin Chen, Jingchang Chen, Jingyang Yuan, Junjie Qiu, Junlong Li, Junxiao Song, Kai Dong, Kai Hu, Kaige Gao, Kang Guan, Kexin Huang, Kuai Yu, Lean Wang, Lecong Zhang, Lei Xu, Leyi Xia, Liang Zhao, Litong Wang, Liyue Zhang, Meng Li, Miaojun Wang, Mingchuan Zhang, Minghua Zhang, Minghui Tang, Mingming Li, Ning Tian, Panpan Huang, Peiyi Wang, Peng Zhang, Qiancheng Wang, Qihao Zhu, Qinyu Chen, Qiushi Du, R.~J. Chen, R.~L. Jin, Ruiqi Ge, Ruisong Zhang, Ruizhe Pan, Runji Wang, Runxin Xu, Ruoyu Zhang, Ruyi Chen, S.~S. Li, Shanghao Lu, Shangyan Zhou, Shanhuang Chen, Shaoqing Wu, Shengfeng Ye, duplicate~removed) (Ye, Shirong Ma, Shiyu Wang, Shuang Zhou, Shuiping Yu, Shunfeng Zhou, Shuting Pan, T.~Wang, Tao Yun, Tian Pei, Tianyu Sun, W.~L. Xiao, Wangding Zeng, Wanjia Zhao, Wei An, Wen Liu, Wenfeng Liang, Wenjun Gao, Wenqin Yu, Wentao Zhang, X.~Q. Li, Xiangyue Jin, Xianzu Wang, Xiao Bi, Xiaodong Liu, Xiaohan Wang, Xiaojin Shen, Xiaokang Chen, Xiaokang Zhang, Xiaosha Chen, Xiaotao Nie, Xiaowen Sun, Xiaoxiang Wang, Xin Cheng, Xin Liu, Xin Xie, Xingchao Liu, Xingkai Yu, Xinnan Song, Xinxia Shan, Xinyi Zhou, Xinyu Yang, Xinyuan Li, Xuecheng Su, Xuheng Lin, Y.~K. Li, Y.~Q. Wang, Y.~X. Wei, Y.~X. Zhu, Yang Zhang, Yanhong Xu, (duplicate removed), Yanping Huang, Yao Li, Yao Zhao, Yaofeng Sun, Yaohui Li, Yaohui Wang, Yi~Yu, Yi~Zheng, Yichao Zhang, Yifan Shi, Yiliang Xiong, Ying He, Ying Tang, Yishi Piao, Yisong Wang, Yixuan Tan, Yiyang Ma, Yiyuan Liu, Yongqiang Guo, Yu~Wu, Yuan Ou, Yuchen Zhu, Yuduan Wang, Yue Gong, Yuheng Zou, Yujia He, Yukun Zha, Yunfan Xiong, Yunxian Ma, Yuting Yan, Yuxiang Luo, Yuxiang You, Yuxuan Liu, Yuyang Zhou, Z.~F. Wu, Z.~Z. Ren, Zehui Ren, Zhangli Sha, Zhe Fu, Zhean Xu, Zhen Huang, Zhen Zhang, Zhenda Xie, Zhengyan Zhang, Zhewen Hao, Zhibin Gou, Zhicheng Ma, Zhigang Yan, Zhihong Shao, Zhipeng Xu, Zhiyu Wu, Zhongyu Zhang, Zhuoshu Li, Zihui Gu, Zijia Zhu, Zijun Liu, Zilin Li, Ziwei Xie, Ziyang Song, Ziyi Gao, and Zizheng Pan.
\newblock {DeepSeek-V3} technical report.
\newblock \emph{{arXiv} preprint arXiv:2412.19437v2}, February 2025.
\newblock URL \url{https://arxiv.org/abs/2412.19437v2}.
\newblock Submitted 18 Feb 2025.

\bibitem[{European Crime Prevention Network (EUCPN)}(2022)]{eucpn2022predictive_policing}
{European Crime Prevention Network (EUCPN)}.
\newblock {Artificial intelligence and predictive policing: risks and challenges}.
\newblock Technical report, European Crime Prevention Network, Brussels, Belgium, 2022.
\newblock Publication 43-2022.

\bibitem[Gemini~Team(2024)]{gemini2024unlocking}
Google Gemini~Team.
\newblock Gemini 1.5: Unlocking multimodal understanding across millions of tokens of context.
\newblock \emph{arXiv preprint arXiv:2403.05530}, 2024.
\newblock URL \url{https://arxiv.org/abs/2403.05530}.

\bibitem[Haller et~al.(2024)Haller, Golde, and Akbik]{haller2024pecc}
Patrick Haller, Jonas Golde, and Alan Akbik.
\newblock {PECC}: Problem extraction and coding challenges.
\newblock In \emph{Proceedings of the 2024 Joint International Conference on Computational Linguistics, Language Resources and Evaluation (LREC-COLING 2024)}, pages 12690--12699, Torino, Italia, May 2024. European Language Resources Association (ELRA) and International Committee on Computational Linguistics (ICCL).
\newblock \doi{10.48550/arXiv.2404.18766}.
\newblock URL \url{https://aclanthology.org/2024.lrec-main.1111}.

\bibitem[Hengle et~al.(2025)Hengle, Bajpai, Dan, and Chakraborty]{hengle2025mlmextended}
Amey Hengle, Prasoon Bajpai, Soham Dan, and Tanmoy Chakraborty.
\newblock Can llms reason over extended multilingual contexts? towards long-context evaluation beyond retrieval and haystacks.
\newblock \emph{arXiv preprint arXiv:2504.12845}, April 2025.
\newblock URL \url{https://arxiv.org/abs/2504.12845}.
\newblock arXiv preprint.

\bibitem[Hsieh et~al.(2024)Hsieh, Sun, Kriman, Acharya, Rekesh, Jia, Zhang, and Ginsburg]{hsieh2024ruler_arxiv}
Cheng-Ping Hsieh, Simeng Sun, Samuel Kriman, Shantanu Acharya, Dima Rekesh, Fei Jia, Yang Zhang, and Boris Ginsburg.
\newblock {RULER: What's the Real Context Size of Your Long-Context Language Models?}
\newblock \emph{{arXiv} preprint arXiv:2404.06654}, 2024.
\newblock URL \url{https://arxiv.org/abs/2404.06654}.

\bibitem[Kamradt(2023)]{kamradt2023needle}
Greg Kamradt.
\newblock Needle in a haystack.
\newblock \url{https://github.com/gkamradt/LLMTest_NeedleInAHaystack}, 2023.
\newblock Accessed: 2025-05-15.

\bibitem[Leidinger et~al.(2024)Leidinger, Van~Rooij, and Shutova]{Leidinger2024LLMsNonmonotonic}
Alina Leidinger, Robert Van~Rooij, and Ekaterina Shutova.
\newblock Are {LLMs} classical or nonmonotonic reasoners? lessons from generics.
\newblock In \emph{Proceedings of the 62nd Annual Meeting of the Association for Computational Linguistics (Volume 2: Short Papers)}, pages 558--573, Bangkok, Thailand, August 2024. Association for Computational Linguistics.
\newblock URL \url{https://aclanthology.org/2024.acl-short.51}.

\bibitem[Li et~al.(2024{\natexlab{a}})Li, Wang, Zheng, and Zhang]{li-etal-2024-loogle}
Jiaqi Li, Mengmeng Wang, Zilong Zheng, and Muhan Zhang.
\newblock {LooGLE}: Can long\-context language models understand long contexts?
\newblock In \emph{Proceedings of the 62nd Annual Meeting of the Association for Computational Linguistics (Volume 1: Long Papers)}, pages 16304--16333, Bangkok, Thailand, August 2024{\natexlab{a}}. Association for Computational Linguistics.
\newblock \doi{10.18653/v1/2024.acl-long.859}.
\newblock URL \url{https://aclanthology.org/2024.acl-long.859/}.

\bibitem[Li et~al.(2025)Li, Zhang, Jiang, Xie, Huang, Wang, and Cheng]{cai2025lara}
Kuan Li, Liwen Zhang, Yong Jiang, Pengjun Xie, Fei Huang, Shuai Wang, and Minhao Cheng.
\newblock Lara: Benchmarking retrieval-augmented generation and long-context llms—no silver bullet for lc or rag routing.
\newblock \emph{arXiv preprint arXiv:2502.09977}, February 2025.
\newblock URL \url{https://arxiv.org/abs/2502.09977}.
\newblock arXiv preprint.

\bibitem[Li et~al.(2024{\natexlab{b}})Li, Zhang, Zhang, Duan, Liu, and Chen]{li_mo2024needlebench}
Mo~Li, Songyang Zhang, Taolin Zhang, Haodong Duan, Yunxin Liu, and Kai Chen.
\newblock {NeedleBench}: {Can} {LLMs} {Do} {Retrieval} and {Reasoning} in {Information-Dense} {Context}?
\newblock \emph{arXiv preprint arXiv:2407.11963}, 2024{\natexlab{b}}.
\newblock Version v2, May 2025. URL: \url{https://doi.org/10.48550/arXiv.2407.11963}.

\bibitem[Ling et~al.(2025)Ling, Liu, Yan, Yang, Lin, Fan, Shen, Du, and Chen]{ling2025longreason}
Zhan Ling, Kang Liu, Kai Yan, Yifan Yang, Weijian Lin, Ting-Han Fan, Lingfeng Shen, Zhengyin Du, and Jiecao Chen.
\newblock {LongReason}: A synthetic long-context reasoning benchmark via context expansion.
\newblock \emph{arXiv preprint arXiv:2501.15089v2}, February 2025.
\newblock URL \url{https://arxiv.org/abs/2501.15089v2}.
\newblock arXiv preprint.

\bibitem[Liu et~al.(2024)Liu, Lin, Hewitt, Paranjape, Bevilacqua, Petroni, and Liang]{liu-etal-2024-modeling}
Nelson~F. Liu, Kevin Lin, John Hewitt, Ashwin Paranjape, Michele Bevilacqua, Fabio Petroni, and Percy Liang.
\newblock {Lost in the Middle: How Language Models Use Long Contexts}.
\newblock \emph{Transactions of the Association for Computational Linguistics}, 12:\penalty0 157--173, 2024.
\newblock URL \url{https://aclanthology.org/2024.tacl-12.157}.

\bibitem[Luo et~al.(2023)Luo, Tan, Huang, Li, and Zhao]{luo2023latent}
Simian Luo, Yiqin Tan, Longbo Huang, Jian Li, and Hang Zhao.
\newblock {Latent Consistency Models}: Synthesizing high-resolution images with few-step inference.
\newblock \emph{{arXiv} preprint arXiv:2310.04378}, 2023.
\newblock URL \url{https://arxiv.org/abs/2310.04378}.

\bibitem[{Meta LCM Team} et~al.(2024){Meta LCM Team}, Barrault, Duquenne, Elbayad, Kozhevnikov, Alastruey, Andrews, Coria, Couairon, Costa-jussà, Dale, Elsahar, Heffernan, Janeiro, Tran, Ropers, Sánchez, San~Roman, Mourachko, Saleem, and Schwenk]{lcm_team2024concept_arxiv}
{Meta LCM Team}, Loïc Barrault, Paul-Ambroise Duquenne, Maha Elbayad, Artyom Kozhevnikov, Belen Alastruey, Pierre Andrews, Mariano Coria, Guillaume Couairon, Marta~R. Costa-jussà, David Dale, Hady Elsahar, Kevin Heffernan, João~Maria Janeiro, Tuan Tran, Christophe Ropers, Eduardo Sánchez, Robin San~Roman, Alexandre Mourachko, Safiyyah Saleem, and Holger Schwenk.
\newblock {Large Concept Models}: Language modeling in a sentence representation space.
\newblock \emph{{arXiv} preprint arXiv:2412.08821}, 2024.
\newblock URL \url{https://arxiv.org/abs/2412.08821}.
\newblock Last revised December 15, 2024.

\bibitem[Mirzadeh et~al.(2024)Mirzadeh, Alizadeh, Shahrokhi, Tuzel, Bengio, and Farajtabar]{mirzadeh2024gsm_symbolic}
Iman Mirzadeh, Keivan Alizadeh, Hooman Shahrokhi, Oncel Tuzel, Samy Bengio, and Mehrdad Farajtabar.
\newblock {GSM-Symbolic}: Understanding the limitations of mathematical reasoning in large language models.
\newblock In \emph{Proceedings of the EACL 2024 Workshop on Mathematical Reasoning}, pages 1--12, Online (Virtual), October 2024. European Chapter of the Association for Computational Linguistics.
\newblock \doi{10.48550/arXiv.2410.05229}.
\newblock URL \url{https://arxiv.org/abs/2410.05229}.

\bibitem[Mondal et~al.(2024)Mondal, Webb, Wang, Krabach, and Momennejad]{mondal2024pfc_planning}
Shanka~Subhra Mondal, Taylor Webb, Chi Wang, Brian Krabach, and Ida Momennejad.
\newblock {A Prefrontal Cortex–Inspired Architecture for Planning in Large Language Models}.
\newblock \emph{{arXiv} preprint arXiv:2310.00194v3}, 2024.
\newblock URL \url{https://arxiv.org/abs/2310.00194v3}.

\bibitem[Muhlgay et~al.(2023)Muhlgay, Ram, Magar, Levine, Ratner, Belinkov, Abend, Leyton-Brown, Shashua, and Shoham]{muhlgay2023factor}
Dor Muhlgay, Ori Ram, Inbal Magar, Yoav Levine, Nir Ratner, Yonatan Belinkov, Omri Abend, Kevin Leyton-Brown, Amnon Shashua, and Yoav Shoham.
\newblock {Generating Benchmarks for Factuality Evaluation of Language Models}.
\newblock \emph{{arXiv} preprint arXiv:2307.06908}, 2023.
\newblock Version v2 (Feb 4 2024). URLs: \url{https://arxiv.org/abs/2307.06908} and \url{https://arxiv.org/pdf/2307.06908v2.pdf}.

\bibitem[Nangia and Bowman(2018)]{nangia2018listops_sereTOD}
Nikita Nangia and Samuel~R. Bowman.
\newblock {ListOps: A Diagnostic Dataset for Latent Tree Learning}.
\newblock \emph{arXiv preprint arXiv:1804.06028v1}, 2018.

\bibitem[{OpenAI}(2025{\natexlab{a}})]{openai2025gpt41_api}
{OpenAI}.
\newblock {GPT-4.1 API Documentation}.
\newblock \url{https://platform.openai.com/docs/models/gpt-4-1}, 2025{\natexlab{a}}.
\newblock Accessed: 2025-05-15.

\bibitem[{OpenAI}(2025{\natexlab{b}})]{openai_pricing_2025}
{OpenAI}.
\newblock {Pricing}.
\newblock \url{https://platform.openai.com/docs/pricing}, 2025{\natexlab{b}}.
\newblock Accessed: 2025-06-05.

\bibitem[Rane et~al.(2024)Rane, Karmalkar, Deshpande, and Vichare]{rane2024blackbox_xai}
Jayesh Rane, Prathamesh Karmalkar, Atharva Deshpande, and Shreyas Vichare.
\newblock {Enhancing black-box models: Advances in explainable artificial intelligence for ethical decision-making}.
\newblock \emph{Multimedia Tools and Applications}, pages 1--26, 2024.
\newblock \doi{10.1007/s11042-023-17879-x}.

\bibitem[Ren et~al.(2024)Ren, Xiao, Qi, Jin, and Wu]{ren2024symtex}
Lin Ren, Guohui Xiao, Guilin Qi, Rihui Jin, and Tongtong Wu.
\newblock {SymTex}: A new benchmark for non\-monotonic reasoning capability of large language models.
\newblock In \emph{Proceedings of the ICLR 2025 Workshop on Reasoning and Planning for Large Language Models}, OpenReview (ICLR 2025 Workshop), September 2024.
\newblock URL \url{https://openreview.net/forum?id=cfDbuobmU0}.
\newblock Submitted 22\,Sep\,2024; Last modified 05\,Feb\,2025.

\bibitem[Shaham et~al.(2023)Shaham, Ivgi, Efrat, Berant, and Levy]{shaham-etal-2023-zeroscrolls}
Uri Shaham, Maor Ivgi, Avia Efrat, Jonathan Berant, and Omer Levy.
\newblock {ZeroSCROLLS}: A zero-shot benchmark for long text understanding.
\newblock In Houda Bouamor, Juan Pino, and Kalika Bali, editors, \emph{Findings of the Association for Computational Linguistics: EMNLP 2023}, pages 7977--7989, Singapore, December 2023. Association for Computational Linguistics.
\newblock \doi{10.18653/v1/2023.findings-emnlp.536}.
\newblock URL \url{https://aclanthology.org/2023.findings-emnlp.536}.

\bibitem[Sheng et~al.(2025)Sheng, Wen, Li, and Zeng]{sheng-etal-2025-evaluating}
Yu~Sheng, Wanting Wen, Linjing Li, and Daniel Zeng.
\newblock {Evaluating Generalization Capability of Language Models across Abductive, Deductive and Inductive Logical Reasoning}.
\newblock In \emph{Proceedings of the 31st International Conference on Computational Linguistics (COLING 2025)}, pages 4945--4957, Abu Dhabi, United Arab Emirates, January 2025. Association for Computational Linguistics.
\newblock URL \url{https://aclanthology.org/2025.coling-main.330}.

\bibitem[Shi et~al.(2023)Shi, Chen, Misra, Scales, Dohan, Chi, Schärli, and Zhou]{shi2023large}
Freda Shi, Xinyun Chen, Kanishka Misra, Nathan Scales, David Dohan, Ed~H. Chi, Nathanael Schärli, and Denny Zhou.
\newblock Large language models can be easily distracted by irrelevant context.
\newblock \emph{arXiv preprint arXiv:2302.00093}, 2023.
\newblock URL \url{https://arxiv.org/abs/2302.00093}.

\bibitem[{Stanford Institute for Human-Centered AI (HAI)}(2025)]{stanford_hai_2025_ai_index}
{Stanford Institute for Human-Centered AI (HAI)}.
\newblock {AI Index Report 2025}.
\newblock Technical report, Stanford University, Stanford, CA, 2025.
\newblock Published annually by the Stanford Institute for Human-Centered AI.

\bibitem[Uzar(2025)]{ContextArena2025}
Dillon Uzar.
\newblock {Context Arena}: Long context llm leaderboard (openai mrcr).
\newblock \url{https://contextarena.ai/}, May 2025.
\newblock Accessed: 2025-06-03. Benchmark inspired by Google DeepMind's MRCR eval (arXiv:2409.12640v2) and uses OpenAI MRCR dataset (openai/mrcr on Hugging Face).

\bibitem[Valmeekam et~al.(2023)Valmeekam, Marquez, Olmo, Sreedharan, and Kambhampati]{valmeekam2023planbench}
Karthik Valmeekam, Matthew Marquez, Alberto Olmo, Sarath Sreedharan, and Subbarao Kambhampati.
\newblock {PlanBench}: An extensible benchmark for evaluating large language models on planning and reasoning about change.
\newblock \emph{Advances in Neural Information Processing Systems}, 36:\penalty0 38975--38987, 2023.

\bibitem[Vishwanath et~al.(2025)Vishwanath, Alyakin, Alber, Lee, Kondziolka, and Oermann]{vishwanath2025meddistractqa}
Krithik Vishwanath, Anton Alyakin, Daniel~Alexander Alber, Jin~Vivian Lee, Douglas Kondziolka, and Eric~Karl Oermann.
\newblock {Medical large language models are easily distracted}.
\newblock \emph{{arXiv} preprint arXiv:2504.01201}, 2025.
\newblock Submitted April 1, 2025. URL: \url{https://arxiv.org/abs/2504.01201}.

\bibitem[Wang et~al.(2024)Wang, Chen, Fu, Liao, Zhang, Wu, Yu, Xu, Zhang, Luo, Li, Yang, Huang, and Li]{wang2024loong_realworld}
Minzheng Wang, Longze Chen, Cheng Fu, Shengyi Liao, Xinghua Zhang, Bingli Wu, Haiyang Yu, Nan Xu, Lei Zhang, Run Luo, Yunshui Li, Min Yang, Fei Huang, and Yongbin Li.
\newblock {Leave No Document Behind: Benchmarking Long-Context LLMs with Extended Multi-Doc QA}.
\newblock \emph{{arXiv} preprint arXiv:2406.17419}, 2024.
\newblock URL \url{https://arxiv.org/abs/2406.17419}.

\bibitem[Wei et~al.(2022)Wei, Wang, Schuurmans, Bosma, Ichter, Xia, Chi, Le, and Zhou]{wei2022chainofthought_neurips}
Jason Wei, Xuezhi Wang, Dale Schuurmans, Maarten Bosma, Brian Ichter, Fei Xia, Ed~H. Chi, Quoc~V. Le, and Denny Zhou.
\newblock {Chain-of-Thought Prompting Elicits Reasoning in Large Language Models}.
\newblock In \emph{Advances in Neural Information Processing Systems}, volume~35, pages 24824--24837, 2022.
\newblock Conference held in 2022; proceedings often list the subsequent year.

\bibitem[Yao et~al.(2023)Yao, Yu, Zhao, Shafran, Griffiths, Cao, and Narasimhan]{yao2023treeofthoughts_neurips}
Shunyu Yao, Dian Yu, Jeffrey Zhao, Izhak Shafran, Thomas~L. Griffiths, Yuan Cao, and Karthik Narasimhan.
\newblock {Tree of Thoughts}: Deliberate problem solving with large language models.
\newblock In \emph{Proceedings of the Thirty‐Seventh Conference on Neural Information Processing Systems}, volume~36, pages 16215--16229, New Orleans, LA, USA, December 2023. Neural Information Processing Systems Foundation.
\newblock \doi{10.5555/3580805.3581569}.
\newblock URL \url{https://proceedings.neurips.cc/paper/2023/hash/271db9922b8d1f4dd7aaef84ed5ac703-Abstract.html}.

\bibitem[Yuan et~al.(2025)Yuan, Li, Feng, Wang, Zhang, Shi, Tan, Pan, Hu, and Li]{Yuan2025Silencer}
Peiwen Yuan, Yiwei Li, Shaoxiong Feng, Xinglin Wang, Yueqi Zhang, Jiayi Shi, Chuyi Tan, Boyuan Pan, Yao Hu, and Kan Li.
\newblock {Silencer}: From discovery to mitigation of self-bias in {LLM-as-Benchmark-Generator}, May 2025.
\newblock URL \url{https://arxiv.org/abs/2505.20738}.

\bibitem[Zhang et~al.(2024)Zhang, Chen, Hu, Xu, Chen, Hao, Han, Thai, Wang, Liu, and Sun]{zhang2024infinitebench_acl}
Xinrong Zhang, Yingfa Chen, Shengding Hu, Zihang Xu, Junhao Chen, Moo~Khai Hao, Xu~Han, Zhen~Leng Thai, Shuo Wang, Zhiyuan Liu, and Maosong Sun.
\newblock {\ensuremath{\infty}Bench: Extending Long Context Evaluation Beyond 100K Tokens}.
\newblock In Lun-Wei Ku, Andre Martins, and Vivek Srikumar, editors, \emph{Proceedings of the 62nd Annual Meeting of the Association for Computational Linguistics (Volume 1: Long Papers)}, pages 15262--15277, Bangkok, Thailand, August 2024. Association for Computational Linguistics.
\newblock \doi{10.18653/v1/2024.acl-long.814}.
\newblock URL \url{https://aclanthology.org/2024.acl-long.814}.

\bibitem[Zuboff(2019)]{zuboff2019surveillance_capitalism}
Shoshana Zuboff.
\newblock \emph{{The Age of Surveillance Capitalism: The Fight for a Human Future at the New Frontier of Power}}.
\newblock PublicAffairs, New York, NY, 2019.

\end{thebibliography}

\clearpage
\appendix
\section*{Appendix}

\section{Extending VLO for Other Reasoning Types / Non-Numberic Reasoning}
\label{sec:extending_VLO}
As VLO is built on a deterministically verifiable Abstract Syntax Tree (AST), this core component can be adapted from numerical operations to symbolic and logical ones. The existing framework can be extended to test abductive, inductive, and defeasible reasoning with non-numeric symbols, while remaining deterministically verifiable.

\subsubsection{Adapting the Core: The Abstract Syntax Tree (AST)}

The current AST uses operators like \texttt{SUM}, \texttt{MAX}, \texttt{MIN} on integer \texttt{Atom} nodes. This can be generalized:

\begin{itemize}
	\item \textbf{Atoms:} The \texttt{Atom} node, which currently holds an integer, can be modified to hold a string representing a non-numeric symbol, a fact, or a concept (e.g., ``the ground is wet'', ``Tweety is a bird'').
	\item \textbf{Operators:} The \texttt{OpNode} can be defined with new operators that represent logical reasoning tasks, such as \texttt{ABDUCE}, \texttt{INDUCE}, or \texttt{DEFEASIBLE\_QUERY}.
\end{itemize}

The crucial step is to define how these new operators are evaluated deterministically in the \texttt{eval\_node} function.

\subsubsection{Making Logical Reasoning Deterministically Verifiable}

The main challenge is making subjective-sounding reasoning tasks verifiable. This is achieved by defining a clear, programmatic evaluation logic for each new operator within a constrained environment.

\paragraph{Abductive Reasoning (Inference to the Best Explanation)}

\begin{itemize}
	\item \textbf{Goal:} To infer the most likely cause given a set of observations.
	\item \textbf{Deterministic Method:} Define a simple, score-based logic. The \texttt{ABDUCE} operator would take a set of observations and a list of potential causes, each with predefined properties. The \texttt{eval\_node} function would calculate a score for each cause and select the ``best'' one based on this score.
	\item \textbf{Example AST:}
	      \begin{verbatim}
(\texttt{ABDUCE}
  (OBSERVATIONS "lights flicker" "strange hum")
  (CAUSES
    (CAUSE "power surge" (simplicity 2) (likelihood 0.8))
    (CAUSE "ghost" (simplicity 8) (likelihood 0.1))
  )
)
\end{verbatim}
	\item \textbf{\texttt{eval\_node} Logic:} It would compute \texttt{score = likelihood / simplicity} for each cause and return the name of the cause with the highest score. In this case, ``power surge'' (0.4) beats ``ghost'' (0.0125). The ground truth is deterministically ``power surge''.
\end{itemize}

\paragraph{Inductive Reasoning (Generalization)}

\begin{itemize}
	\item \textbf{Goal:} To form a general rule from specific examples.
	\item \textbf{Deterministic Method:} Constrain the space of possible rules. The \texttt{INDUCE} operator would take a list of examples and a predefined set of potential rules. The \texttt{eval\_node} function would select the first rule from the set that is consistent with all provided examples.
	\item \textbf{Example AST:}
	      \begin{verbatim}
(\texttt{INDUCE}
  (EXAMPLES ("raven A is black") ("raven B is black"))
  (RULE_CANDIDATES ("all birds are black") ("all ravens are black") 
                   ("some ravens are black"))
)
\end{verbatim}
	\item \textbf{\texttt{eval\_node} Logic:} It would check each rule candidate. ``all birds are black'' is not contradicted but is less specific. ``all ravens are black'' is consistent. It would return ``all ravens are black'' as the correct, most specific, consistent rule from the given set.
\end{itemize}

\paragraph{Defeasible Reasoning (Rules with Exceptions)}

\begin{itemize}
	\item \textbf{Goal:} To reason with rules that can be defeated by new information.
	\item \textbf{Deterministic Method:} Implement a simple, priority-based logic system. The \texttt{DEFEASIBLE\_QUERY} operator would take a set of facts and an ordered list of rules and exceptions. The \texttt{eval\_node} function applies the rules in order, allowing later rules (exceptions) to override earlier ones.
	\item \textbf{Example AST:}
	      \begin{verbatim}
(\texttt{DEFEASIBLE\_QUERY}
  (QUERY "Tweety can fly")
  (KNOWLEDGE_BASE
    (FACT "Tweety is a bird")
    (FACT "Tweety is a penguin")
    (RULE "all birds can fly" (priority 1))
    (EXCEPTION "penguins cannot fly" (priority 2))
  )
)
\end{verbatim}
	\item \textbf{\texttt{eval\_node} Logic:} It would first conclude ``Tweety can fly'' from the priority 1 rule. Then, it would process the priority 2 exception, which defeats the initial conclusion. The final, deterministic ground truth is \texttt{False} (i.e., ``Tweety cannot fly'').
\end{itemize}

\subsubsection{Extending the Generation and Validation Pipeline}

With a deterministic AST in place, the rest of the \texttt{verbose-ListOps.py} pipeline can be adapted:

\begin{enumerate}
	\item \textbf{\texttt{build\_random\_ast}:} This function would be updated to construct these new symbolic and logical ASTs from a set of predefined templates to ensure the generated problems coherence.

	\item \textbf{\texttt{generate\_narrative}:}
	      \begin{itemize}
		      \item \textbf{Prompts:} The prompts would be modified. Instead of asking the LLM to narrate a scene about finding the \texttt{MAX} of a set of numbers, you would ask it to narrate a scene where characters reason about the most likely cause of an event.
		      \item \textbf{Narrative Anchors:} The concept of ``narrative anchors'' is even more powerful here. The result of an \texttt{ABDUCE} operation (``power surge'') could be given the anchor ``The Prime Theory,'' which then becomes a symbolic input for a subsequent reasoning step.
	      \end{itemize}

	\item \textbf{Validation (\texttt{make\_number\_validator} and \texttt{validator.py}):} This part requires the most significant rewrite, shifting from numerical validation to symbolic validation.
	      \begin{itemize}
		      \item \textbf{The Goal Remains:} The core validation goals are the same: ensure all required inputs are mentioned, the (now symbolic) result is kept implicit, and no extraneous information or conclusions are leaked.
		      \item \textbf{New Logic:} Instead of \texttt{extract\_numbers\_from\_text}, a function like \texttt{extract\_facts\_from\_text} that uses string matching or regex to verify that the narrative mentions would need be added, for example, ``the lights flicker'' and ``a strange hum''.
		      \item \textbf{Implicit Result:} The validator would check that the word ``power surge'' is \textit{not} explicitly stated as the conclusion, but is only referenced by its anchor (``The Prime Theory'').
	      \end{itemize}
\end{enumerate}

\subsubsection{Summary}

\texttt{verbose-ListOps.py}'s modular, AST-driven, and agentic-validation architecture provides a robust and sophisticated foundation to extend VLO to test deeper, symbolic reasoning by:

\begin{enumerate}
	\item \textbf{Defining the symbolic operators} and their deterministic evaluation logic.
	\item \textbf{Creating templates for generating coherent, symbolic ASTs.}
	\item \textbf{Rewriting the prompt templates} in \texttt{\_generate\_narrative\_recursive} to guide the LLM in narrating these logical problems.
	\item \textbf{Replacing the numerical validation logic} with a symbolic/factual validation system that enforces the same core principles of operand presence and result implicitness.
\end{enumerate}

The result would be a novel and powerful benchmark that pushes LLMs beyond numerical computation into the realm of structured, verifiable, narrative-based logical reasoning.

\section{Models Evaluated}
\label{app:models_evaluated}
Table \ref{tab:appendix_models_evaluated} lists the Large Language Models (LLMs) evaluated in this study on the Verbose ListOps benchmark. These models represent a range of state-of-the-art closed-source and open-source offerings available at the time of evaluation (16 May 2025).

\begin{table}[h!]
	\centering
	\caption{Large Language Models Evaluated on Verbose ListOps.}
	\label{tab:appendix_models_evaluated}
	\resizebox{\textwidth}{!}{%
		\begin{tabular}{llll}
			\toprule
			\textbf{Model Name}    & \textbf{Version}                & \textbf{Max Input Tok.}  & \textbf{OpenRouter Identifier}             \\
			\midrule
			\multicolumn{4}{l}{\textit{Closed-Source Models}}                                                                                \\
			Gemini 2.5 Pro         & \texttt{preview-05-06}          & \SI{1}{M}                & \texttt{gemini-2.5-pro-preview}            \\
			Gemini 2.5 Flash       & \texttt{preview-04-17}          & \SI{1}{M}                & \texttt{gemini-2.5-flash-preview:thinking} \\
			o4 Mini High           & \texttt{2024-04-16}             & \SI{128}{K}              & \texttt{o4-mini-high}                      \\
			GPT-4.1                & \texttt{2025-04-14}             & \SI{1}{M}                & \texttt{gpt-4.1}                           \\
			Claude 3.7 Sonnet High & \texttt{20250219}               & \SI{200}{K}+             & \texttt{claude-3.7-sonnet:thinking}        \\
			Grok 3 Mini High       & \texttt{latest (tested 16 May)} & \SI{128}{K}              & \texttt{grok-3-mini-beta}                  \\
			\midrule
			\multicolumn{4}{l}{\textit{Open-Source Models}}                                                                                  \\
			DeepSeek R1            & \texttt{2025/01/20}             & Varies                   & \texttt{deepseek-r1}                       \\
			DeepSeek V3            & \texttt{0324}                   & Varies                   & \texttt{deepseek-chat-v3-0324}             \\
			Qwen 3 235B A22B       & \texttt{latest (tested 16 May)} & \SI{128}{K}+             & \texttt{qwen3-235b-a22b}                   \\
			Llama 4 Maverick       & \texttt{Apr 5 2025}             & \SI{128}{K} - \SI{10}{M} & \texttt{llama-4-maverick}                  \\
			\bottomrule
		\end{tabular}%
	}
	\vspace{5pt}
	\scriptsize{\textit{Note:} Claimed max input tokens are approximate and subject to change based on provider updates. Identifiers are illustrative examples based on common API/HF naming conventions and may vary. The specific models used for evaluation are as listed in Table 2 of the main paper.}
\end{table}

\section{Verbose ListOps Generation Hyperparameters}
\label{app:generation_hyperparams}
The Verbose ListOps benchmark instances were programmatically generated. Key parameters and settings for the generation process are detailed below. These correspond to the \texttt{Config} dataclass and other settings in the \texttt{verbose-ListOps.py} script.

\subsection{Core ListOps Parameters}
\begin{itemize}
	\item \textbf{Maximum Operations (\texttt{MAX\_OPS}):} 8 (Controls the maximum depth/complexity of the ListOps Abstract Syntax Tree).
	\item \textbf{Maximum Branching Factor (\texttt{MAX\_BRANCH}):} 8 (Maximum number of children for any operation node).
	\item \textbf{Minimum Arity (\texttt{MIN\_ARITY}):} 4 (Minimum number of children for any operation node).
	\item \textbf{Atom Value Range (\texttt{MIN\_ATOM\_VAL}, \texttt{MAX\_ATOM\_VAL}):} 1 to 30.
	\item \textbf{Early Termination Probability (\texttt{EARLY\_TERMINATION\_PROBABILITY}):} 0.0 (Probability of terminating Abstract Syntax Tree branch growth before \texttt{MAX\_OPS} is reached).
\end{itemize}

\subsection{Narrative Generation Parameters}
\begin{itemize}
	\item \textbf{Narrative Generation Large Language Model (\texttt{MODEL} in \texttt{verbose-ListOps.py}):} \texttt{google/gemini-2.5-flash-preview:thinking} (accessed via OpenRouter API).
	\item \textbf{Target Total Tokens (\texttt{MAX\_TOTAL\_TOKENS}):} \num{10000} (For the experiments reported in Table 2). The script can generate longer narratives.
	\item \textbf{Padding Max Token Percentage (\texttt{PADDING\_MAX\_TOK\_PERCENT}):} 0.75 (Maximum percentage of the *remaining* token budget (after beats) that can be used for padding).
	\item \textbf{Max Padding Paragraphs per Slot (\texttt{MAX\_PAD\_PARAGRAPHS}):} 30.
	\item \textbf{Use Narrative Anchors (\texttt{USE\_NARRATIVE\_ANCHORS}):} True (Conceptual names for intermediate results).
	\item \textbf{Use Large Language Model for Anchor Naming (\texttt{USE\_LLM\_NAMING}):} True.
	\item \textbf{World Generation Parameters:}
	      \begin{itemize}
		      \item Min/Max Characters (\texttt{MIN\_WORLD\_CHARS}, \texttt{MAX\_WORLD\_CHARS}): 6 to 8.
		      \item Min/Max Concepts (\texttt{MIN\_WORLD\_CONCEPTS}, \texttt{MAX\_WORLD\_CONCEPTS}): 3 to 7.
		      \item World Generation Temperature (\texttt{WORLD\_GEN\_TEMP}): 0.9.
	      \end{itemize}
	\item \textbf{Beat Generation Temperature (\texttt{BEAT\_GEN\_TEMP}):} 0.5.
	\item \textbf{Creative Narrative Temperature (for Intro/Padding) (\texttt{CREATIVE\_NARRATIVE\_TEMP}):} 0.5.
	\item \textbf{Anchor Generation Temperature (\texttt{ANCHOR\_GEN\_TEMP}):} 0.85.
\end{itemize}

\subsection{Iterative Validation and Retry Parameters}
\begin{itemize}
	\item \textbf{Large Language Model Validator Model (\texttt{LLM\_VALIDATOR\_MODEL}):} \texttt{google/gemini-2.5-flash-preview:thinking} (Used in the iterative beat generation loop).
	\item \textbf{Large Language Model Validator Temperature (\texttt{LLM\_VALIDATOR\_TEMP}):} 0.05.
	\item \textbf{Beat Revision Temperature (\texttt{BEAT\_REVISION\_TEMP}):} 0.1.
	\item \textbf{Max Large Language Model Validation Iterations (\texttt{MAX\_LLM\_VALIDATION\_ITERATIONS}):} 6 (Internal loop for a single beat).
	\item \textbf{Max Beat Retries (Outer Loop) (\texttt{MAX\_BEAT\_RETRIES}):} 5.
	\item \textbf{Max Padding Retries (\texttt{MAX\_PAD\_RETRIES}):} 7.
	\item \textbf{Max Introduction Scene Retries (\texttt{INTRO\_MAX\_RETRIES}):} 3.
	\item \textbf{Max World Generation Retries (\texttt{WORLDGEN\_MAX\_RETRIES}):} 5.
	\item \textbf{Retry Initial Delay (\texttt{RETRY\_INITIAL\_DELAY}):} 0.25 seconds (for general API call retries).
\end{itemize}

\subsection{Token and API Settings}
\begin{itemize}
	\item \textbf{Tokenizer (\texttt{encoder}):} \texttt{cl100k\_base} (via \texttt{tiktoken}).
	\item \textbf{Max API Token Limit (\texttt{MAX\_API\_TOKEN\_LIMIT}):} \num{60000} (Safety buffer for Large Language Model calls, allowing space for reasoning tokens if supported by the model endpoint).
	\item \textbf{Max Tokens Buffer (\texttt{MAX\_TOKENS\_BUFFER}):} 500 (Safety margin when checking against \texttt{MAX\_TOTAL\_TOKENS}).
	\item \textbf{Max Requests Per Second (\texttt{MAX\_REQUESTS\_PER\_SECOND}):} 900.0 (Target for OpenRouter rate limiter, dynamically adjusted).
\end{itemize}

\subsection{Generation Cost}
The generation of the 1000 samples (each $\approx$\SI{10}{k} tokens) for the main evaluation incurred an estimated API cost of approximately \$1500 USD using the OpenRouter API with the specified generation and validator Large Language Models. Evaluating all listed models on these samples incurred an additional estimated API cost of approximately \$500 USD.

\section{Dataset Generation Pipeline}
\label{app:generation_pipeline}
The Verbose ListOps dataset is generated programmatically using an agentic pipeline orchestrated by the \texttt{verbose-ListOps.py} script. The process for each sample involves:
\begin{enumerate}
	\item \textbf{Abstract Syntax Tree Generation:} A random ListOps Abstract Syntax Tree (AST) is constructed based on the core ListOps parameters (Appendix \ref{app:generation_hyperparams}). The Abstract Syntax Tree is then evaluated to determine the ground truth answer.
	\item \textbf{World Generation:} An Large Language Model (Gemini 2.5 Flash) generates fictional world metadata (characters, genre, setting, primary object) based on a structured prompt and schema (see Appendix \ref{app:prompts_worldgen}).
	\item \textbf{Narrative Anchor Generation:} If \texttt{USE\_NARRATIVE\_ANCHORS} is true, conceptual names (anchors) for the results of each operation node in the Abstract Syntax Tree are generated, either by an Large Language Model or deterministically.
	\item \textbf{Introduction Scene Generation:} An introductory scene is generated by the Large Language Model, setting the stage without revealing numerical details. This scene is validated for numerical compliance (strict zero numbers, with minor exceptions for phrasing).
	\item \textbf{Iterative Beat Generation and Validation:}
	      \begin{itemize}
		      \item The Abstract Syntax Tree is traversed in post-order. For each \texttt{OpNode}, a narrative "beat" is generated.
		      \item The Large Language Model generator is provided with a detailed prompt including the current operation, conceptual inputs (anchors from child nodes), new atomic inputs, and an extensive set of "ultra-strict number rules" (see Appendix \ref{app:prompts_beatgen}). These rules enforce that only current atomic operands are stated numerically, prior results are referenced by anchors, and the current operation's result is implied.
		      \item The generated beat undergoes an iterative validation loop (\texttt{\_generate\_and\_llm\_validate\_beat} function):
		            \begin{enumerate}
			            \item The beat is first validated by another Large Language Model call (\texttt{LLM\_VALIDATOR\_MODEL}) against the strict rules, using a structured JSON schema for the validator's response.
			            \item If the Large Language Model validator fails the beat, its feedback is used to prompt the generator Large Language Model for a revision. This loop continues for up to \texttt{MAX\_LLM\_VALIDATION\_ITERATIONS}.
		            \end{enumerate}
		      \item If a beat passes the internal Large Language Model validation loop, it is then subjected to a final Python-based programmatic validation (\texttt{make\_number\_validator}) to ensure precise numerical compliance (correct numbers mentioned with exact frequencies, no forbidden numbers, result implicitness).
		      \item If a beat fails either the iterative Large Language Model validation or the final Python validation after all retries (\texttt{MAX\_BEAT\_RETRIES} for the outer loop), the generation for that entire sample is aborted.
	      \end{itemize}
	\item \textbf{Padding Generation:} Between valid beats (except after the root node's beat), optional narrative padding can be inserted to increase context length. Padding is also Large Language Model-generated and validated for numerical compliance (strict zero numbers).
	\item \textbf{Final Question Assembly:} A question asking for the final result of the ListOps sequence is appended to the narrative.
	\item \textbf{Output Formatting:} Successfully generated samples are saved in JSONL format, including the full narrative, Abstract Syntax Tree, ground truth, and metadata.
\end{enumerate}
The generation process utilizes a \texttt{ThreadPoolExecutor} for parallel generation of multiple samples, with up to \texttt{DEFAULT\_MAX\_WORKERS} (100 by default). API calls to OpenRouter are managed by a rate limiter.

\section{Prompts}
\label{app:prompts}
This section provides examples of key prompts used in the Verbose ListOps generation pipeline. Note that these are templates and are dynamically filled with specific details (world information, Abstract Syntax Tree node data, etc.) at runtime.

\subsection{World Generation Prompt}
\label{app:prompts_worldgen}
The Large Language Model is prompted to generate fictional world metadata (characters, genre, setting, object) in a structured JSON format.
\begin{verbatim}
System: You are an expert system designed to generate structured data in
**strictly valid JSON format**. Your task is to create fictional world metadata.
**CRITICAL JSON FORMATTING RULES (MUST FOLLOW EXACTLY):**
1.  The entire output MUST be a single, valid JSON object.
2.  All string keys and string values within the JSON must be enclosed in
    double quotes (e.g., "name": "value").
3.  **If a string value itself needs to contain a double quote character
    (e.g., a nickname within a name), that internal double quote MUST be
    escaped with a backslash (`\\`)**. For example, if a character's name is
    `Dr. "Nickname" Who`, it must be represented in the JSON string as
    `"name": "Dr. \\"Nickname\\" Who"`.
4.  Ensure all commas, colons, curly braces `{{}}`, and square brackets `[]`
    are correctly placed according to standard JSON syntax.
5.  Do not include any text, explanations, or markdown (like ```json)
    before or after the single JSON object.

**Instructions for Content Generation:**
1.  **Characters:** Generate exactly {num_characters} distinct characters. Each...
    *   `name`: string (e.g., "Kaelen Vane", "Seraphina Moonwhisper")
    *   `role`: string (e.g., "The grizzled warrior," "The cunning sorceress,")
    *   `quirk`: string (e.g., "Collects antique spoons," "Only speaks in riddles,")
2.  **Genre:** Define a `genre` as a string (e.g., "Steampunk Adventure").
3.  **Setting:** Define a `setting` as a string (e.g., "A floating city...").
4.  **Object:** Define an `object` as a string (plural noun, e.g., "etherium crystals").

**Guidance for Content:** Strive for thematic coherence...
Output ONLY the single, valid JSON object.

User: (Dynamically filled with num_characters)
\end{verbatim}
The full prompt includes detailed examples and constraints for each field, ensuring the output adheres to the \texttt{WORLD\_SCHEMA}.

\subsection{Narrative Anchor Generation Prompt}
\label{app:prompts_anchor}
For \texttt{OpNode}s, conceptual names (anchors) are generated to refer to their results.
\begin{verbatim}
System: You are a master {genre} storyteller and creative naming expert.
Your task is to generate a short, evocative, and thematic 'narrative anchor'.
A narrative anchor is a creative, conceptual name that serves as a descriptive
**label** or **stand-in** for the *result* of a specific event or calculation.

Key Guidelines:
1.  **Thematic:** MUST fit Genre, Setting, Primary Object.
2.  **Concise:** 2 to {MAX_ANCHOR_WORDS} words (e.g., 'The Sunstone's Core').
3.  **No Numbers:** Absolutely no numerical values.
4.  **No Direct Math Terms:** Avoid '\texttt{SUM}', '\texttt{MIN}', '\texttt{MAX}', etc.
5.  **Represent Outcome:** Conceptually represent the result.
6.  **Focus on Noun:** Should feel like a "thing" or "state".
7.  **ABSOLUTE UNIQUENESS:** MUST NOT be in 'List of anchors ALREADY USED'.
    If unable, respond with "UNIQUE_FAILURE".

User:
Genre: {genre}
Setting: {setting}
Item: {primary_object}
Concept/Operation Hint: {concept_keywords_for_prompt}

**List of anchors ALREADY USED...:**
{all_previous_anchors}

Provide ONLY the new, unique anchor name... or 'UNIQUE_FAILURE'.
\end{verbatim}

\subsection{Introduction Scene Generation Prompt}
\label{app:prompts_intro}
The introductory scene sets the stage.
\begin{verbatim}
System: You are a master {genre} storyteller. Your task is to write a
compelling introductory scene... establish setting, introduce characters,
hint at mystery related to {primary_object}.

**ABSOLUTE NUMERICAL RULE FOR THIS INTRODUCTORY SCENE (CRITICAL):**
1.  **ZERO NUMBERS IS THE PRIMARY GOAL:** Use NO numerical values (digits or words).
2.  **EXTREMELY LIMITED EXCEPTION:** MAY use 'one', 'two', or 'three' for
    general, non-quantitative phrasing IF UNAVOIDABLE. NO OTHER NUMBERS.
3.  **HANDLING CHARACTER NAMES WITH DIGITS:** Avoid stating numerical part as quantity.
    Safer to avoid names with digits for intro.
Focus on atmosphere, intrigue... Output ONLY the narrative text.

User:
**World Context:**
- Genre: {genre}
- Setting: {setting}
- Primary Object of Interest: {object}
- Characters to potentially feature: {char_names_roles}

**Task:** Write an engaging introductory scene...
**CRITICAL REMINDER - ADHERE TO THE ABSOLUTE NUMERICAL RULE...**
Output ONLY the narrative text.
\end{verbatim}

\subsection{Main Beat Generation Prompt (Illustrative Core Rules)}
\label{app:prompts_beatgen}
This is the most complex prompt, dynamically constructed for each \texttt{OpNode}. The core is the \texttt{ultra\_strict\_instruction} section. Below is a conceptual summary of its key components:

\begin{verbatim}
System: You are a master {genre} storyteller with an exceptional eye for detail...
Your paramount responsibilities for this scene are:
1.  **Narrative Coherence:** ...
2.  **ULTRA-STRICT NUMERICAL AND OPERATIONAL PRECISION:** ...
    *   **Rule 1.A (Exact Atomic Frequencies):** Mention EACH required *new atomic*
        number EXACTLY the specified number of times... AVOID summarizing.
    *   **Rule 1.C (Conceptual Inputs):** Ensure prior results (conceptual inputs)
        are *active numerical inputs* to THIS scene's operation.
    *   **Operational Fidelity:** Narrated action MUST accurately reflect the
        mathematical operation on ALL inputs.
    *   **Rule 2 (Implicit Outcome):** Numerical result of THIS scene's operation MUST NOT
        be stated explicitly.
    *   **Rule 4 & 5 (Forbidden & No Other Numbers):** ...
Output ONLY the clean narrative text...

User:
Story Scene Task: Create the narrative for the step resulting in '{current_node_conceptual_name}'
(Scene {current_beat_num}/{total_beats})

**Background for Your Scene...:**
- Genre: {world_genre}
- Setting: {world_setting}
- Central Items: {primary_object_as_string}
- Quantities from Previous Events (Conceptual Names & values for YOUR understanding -
  DO NOT use values in story, DO use names if Rule 1.C applies): {conceptual_inputs_context_str}
- New Numbers Introduced (Values & required frequencies for YOUR understanding - Use word
  form, ALL must be mentioned with exact frequencies as per Rule 1.A): {atomic_inputs_context_str_detailed_for_prompt}

**Your Scene's Core Action & Narrative Goal (Follow this closely):**
This scene needs to narrate an event or discovery that mirrors the mathematical
operation: **{op_label}**. The central items are '{safe_primary_object_for_fstring}'.
Inputs to consider:
  1. **Conceptual Inputs:** {conceptual_input_names_only_str_for_action}.
  2. **New Atomic Number Inputs:** {atomic_inputs_context_str_detailed_for_prompt}.
Your narrative must clearly show ALL these inputs... being involved in an action
that reflects the '{op_label}' operation.
  - **Action (Specific to Op, e.g., \texttt{SUM}):** Characters combine/tally ALL inputs...
The outcome will be conceptually known as '{current_node_conceptual_name}'.
Its actual numerical size ('{num_to_words(correct_result)}') must not be stated.

**Narrative Challenge & Your Writing Guide for This Scene (CRITICAL...):**
**1. Key Details to Feature (Inputs in Action & Their EXACT Frequencies):**

[...refer to codebase for full prompt...]

{prior_results_handling_rule_for_prompt} (This is Rule 6)

**Operational Fidelity (CRITICAL):** The narrated action MUST accurately reflect
the mathematical operation '{op_label}' on ALL inputs (conceptual & atomic).
...
**MANDATORY PRE-WRITING CHECKLIST & MENTAL WALKTHROUGH...:**
(Detailed checklist items for the LLM to mentally verify its plan against each rule)
...
**Continue From (End of last scene):**
"...{context_snippet}..."

**Your Response:**
Write ONLY the narrative text for this new scene...
\end{verbatim}
The actual prompt is highly detailed, including specific examples for \texttt{MEDIAN} operations and a pre-writing checklist for the Large Language Model. The \texttt{ultra\_strict\_instruction} section is dynamically built based on the current node's operation, its inputs (atomic and conceptual), its result, and the overall Abstract Syntax Tree context to define precisely which numbers are allowed, required (with exact frequencies), or forbidden for that specific beat.

\subsection{Large Language Model Validator Prompt (Iterative Beat Validation)}
\label{app:prompts_llm_validator}
During the iterative beat generation, another Large Language Model validates the generator's output.
\begin{verbatim}
System: You are an AI numerical compliance checker and literary critic.
Your ONLY task is to evaluate a story 'beat' against a provided set of
ULTRA-STRICT numerical and storytelling rules.
You MUST output your response as a single, valid JSON object and NOTHING ELSE,
adhering precisely to the provided schema.
Your analysis must be meticulous, focusing on exact numerical frequencies...

User:
You are an AI numerical compliance checker. Evaluate the 'Generated Beat Text'
below with ABSOLUTE PRECISION regarding its numerical content, operational
fidelity, and adherence to the 'ULTRA-STRICT NUMBER RULES (Generator's
Writing Guide)' provided.

[...see codebase for full prompt...]

**ULTRA-STRICT NUMBER RULES (Generator's Writing Guide - GROUND TRUTH FOR VALIDATION):**
---
{ultra_strict_instruction_for_llm_validator_context} (This is the full ruleset given to the generator)
---

**VALIDATION ALGORITHM - FOLLOW EXACTLY...:**
(Detailed step-by-step algorithm for the validator LLM to check each rule:
 Phase 1: Number Identification & Counting.
 Phase 2: Rule-by-Rule Compliance Check (Rule 0.A Conceptual Inputs,
          0.C Operational Fidelity, 1.A Atomic Frequencies, 1.B No Re-listing,
          2 Outcome Handling, 3 Permitted Flourishes, 4 Forbidden, 5 No Other,
          6 Prior Result Handling).
 Phase 3: Constructing JSON Response according to VALIDATOR_RESPONSE_SCHEMA,
          including `is_valid`, `explanation_for_generator`, `explanation_for_audit`,
          `overall_revision_summary_for_generator_prompt`, `suggested_revisions`.)

**Generated Beat Text to Evaluate:**
---
{generated_text_cleaned}
---
\end{verbatim}
The validator's prompt includes the full set of rules given to the generator, the generated text, and a detailed algorithm for how the validator should check compliance and structure its JSON response.

\subsection{Final Question Template}
\label{app:prompts_final_question}
The template for the final question appended to the narrative.
\begin{verbatim}
FINAL_QUESTION_TEMPLATE = Template(
    "\n\n---\n\n**Question:** The story describes a sequence of operations that "
    "modify a quantifiable measure related to '$primary_object'. Following this "
    "entire sequence, what is the final, precise numerical value of this measure "
    "at the conclusion of all activities? Provide only the single integer."
)
\end{verbatim}
Here, \texttt{\$primary\_object} is substituted with the specific object generated for the world (e.g., "etherium crystals").

\section{Dataset Details and Access}
\label{app:dataset_details}
The Verbose ListOps dataset is available on Hugging Face Datasets:
\begin{itemize}
	\item \textbf{Dataset Link:} \url{https://huggingface.co/datasets/NeurIPSDB2025-shj32df/verbose-ListOps}
	\item \textbf{Croissant Metadata:} A Croissant metadata file for enhanced discoverability and interoperability is available at: \url{https://huggingface.co/api/datasets/NeurIPSDB2025-shj32df/verbose-ListOps/croissant}
\end{itemize}

The dataset is provided in JSON Lines (\texttt{.jsonl}) format.
\begin{itemize}
	\item \texttt{[1\_RESEARCHER\_DETAIL]\_DATASET\_\*.jsonl}: Contains comprehensive raw generated data for each sample, including detailed generation metadata, full Abstract Syntax Tree structure, all scenes, conceptual references, and beat revision logs. This file is primarily for research and debugging the generation process.
	\item \texttt{[2\_EVAL\_READY]\_DATASET\_\*.jsonl}: A leaner version containing all successfully generated samples by \texttt{verbose-ListOps.py}, with fields relevant for model evaluation. This is the dataset before external validation by \texttt{validator.py}.
	\item \texttt{[4\_FINAL\_EVAL\_CLEANED]\_DATASET\_\*.jsonl}: The final, cleaned dataset intended for benchmarking. This version contains only samples that have passed an additional validation step by the \texttt{validator.py} script, ensuring higher fidelity of the narrative to the underlying ListOps task according to an external Large Language Model judge.
\end{itemize}
The specific dataset used for the results reported in this paper is the \texttt{[4\_FINAL\_EVAL\_CLEANED]} version corresponding to the \SI{10}{k} token length narratives.

\subsection{Key Dataset Fields (in \texttt{EVAL\_READY} and \texttt{FINAL\_EVAL\_CLEANED} versions)}
Table \ref{tab:appendix_dataset_fields} describes the main fields in the evaluation-ready JSONL files.
\begin{table}[h!]
	\centering
	\caption{Key fields in the Verbose ListOps evaluation-ready dataset files.}
	\label{tab:appendix_dataset_fields}
	\begin{tabular}{p{0.25\linewidth} p{0.65\linewidth}}
		\toprule
		\textbf{Field Name}              & \textbf{Description}                                                                                                                             \\
		\midrule
		\texttt{id}                      & Unique identifier for the sample (string).                                                                                                       \\
		\texttt{full\_text\_for\_eval}   & The complete text provided to the Large Language Model for evaluation, consisting of the narrative body followed by the final question (string). \\
		\texttt{ground\_truth\_value}    & The single integer answer to the ListOps problem (integer).                                                                                      \\
		\texttt{ast\_str}                & A string representation of the ListOps Abstract Syntax Tree in prefix notation (string).                                                         \\
		\texttt{num\_operations}         & The total number of operation nodes (non-atom nodes) in the Abstract Syntax Tree (integer).                                                      \\
		\texttt{token\_count\_narrative} & The approximate token count of the narrative body (excluding the question), based on \texttt{cl100k\_base} tokenizer (integer).                  \\
		\bottomrule
	\end{tabular}
\end{table}
The \texttt{RESEARCHER\_DETAIL} file contains additional fields like \texttt{world\_data}, \texttt{scenes\_detail}, \texttt{conceptual\_references}, \texttt{beat\_revision\_details}, and extensive \texttt{generation\_metadata}.

\section{Evaluation Details}
\label{app:evaluation_details}
\begin{itemize}
	\item \textbf{Evaluation Metric:} Performance is measured by exact match accuracy. The Large Language Model's predicted single integer answer must exactly match the \texttt{ground\_truth\_value} for the sample.
	\item \textbf{Number of Samples:} For the main results reported in Table 2, each model was evaluated on 1000 distinct Verbose ListOps samples (from the \texttt{FINAL\_EVAL\_CLEANED} dataset version, with $\approx$\SI{10}{k} token narrative length).
	\item \textbf{External Validation (\texttt{validator.py}):} The \texttt{validator.py} script performs an additional layer of validation on the generated narratives. It uses a separate, more powerful thinking Large Language Model (\texttt{google/gemini-2.5-pro-preview} as specified in \texttt{validator.py}) with extensive prompting to assess whether each step in the narrative correctly reflects the corresponding Abstract Syntax Tree operation, its inputs, and implied result, according to the benchmark's rules (including implicit intermediate results and conceptual referencing). Samples that fail this external validation are excluded from the \texttt{FINAL\_EVAL\_CLEANED} dataset. This script outputs detailed validation results and helps ensure the quality and fidelity of the benchmark instances used for final model evaluation.
\end{itemize}

\section{Code Availability}
\label{app:code_availability}
The code for generating the Verbose ListOps benchmark and the \texttt{validator.py} script are open-sourced and available at:
\begin{itemize}
	\item \textbf{GitHub Repository:} \url{https://github.com/Neurips-anon-h1ndi29v/verbose-ListOps}.
	\item \textbf{Dataset:} \url{https://huggingface.co/datasets/NeurIPSDB2025-shj32df/verbose-ListOps/tree/main}.
\end{itemize}
The repository contains the \texttt{verbose-ListOps.py} script for dataset generation and the \texttt{validator.py} script for post-generation validation and cleaning. Detailed instructions for running the scripts and reproducing the dataset are provided in the repository's README file.

\end{document}